\documentclass[11pt]{article}

\usepackage[T1]{fontenc}
\usepackage{lmodern}
\usepackage[margin=1in]{geometry}
\usepackage[authoryear]{natbib}
\usepackage{amsmath,amssymb}
\usepackage{booktabs,tabularx,array,multirow,float,longtable}
\usepackage{graphicx}
\usepackage{xcolor}
\usepackage{enumitem}
\usepackage{microtype}
\usepackage{siunitx}
\usepackage{hyperref}

\hypersetup{
  hidelinks,
  pdftitle={Asymmetric Bidirectional Predictive Dynamics in Google Search and COVID-19 Time Series: A Multi-Horizon Forecast-Error Analysis},
  pdfauthor={Andrei Velichko and Imen Mahmoud}
}
\newcommand{\FEDE}{\mathrm{FEDE}}
\newcommand{\RMSE}{\mathrm{RMSE}}
\newcolumntype{Y}{>{\raggedright\arraybackslash}X}
\title{Asymmetric Bidirectional Predictive Dynamics in Google Search and COVID-19 Time Series: A Multi-Horizon Forecast-Error Analysis}
\author{%
Andrei Velichko$^{1,*}$ \qquad Imen Mahmoud$^{2}$\\[0.6em]
\small $^{1}$Institute of Physics and Technology, Petrozavodsk State University,\\
\small 33 Lenina Prospekt, Petrozavodsk 185910, Republic of Karelia, Russia\\
\small $^{2}$University of Mannouba, Tunisia\\
\small Laboratory of Research in Innovative and Management, Risk, Accounting and Finance\\[0.4em]
\small $^{*}$Corresponding author: \texttt{velichkogf@gmail.com}\\
\small ORCID: A. Velichko 0000-0002-9341-1831; I. Mahmoud 0009-0006-3686-7249
}
\date{}

\begin{document}
\maketitle

\begin{abstract}
Online search activity is widely used as a digital trace of information seeking during epidemics, but its predictive value is often reduced to correlation, a selected lag, or a single forecast-error score. We examine eight pre-specified Google Trends--COVID-19 pairs in both directions using matched LightGBM forecasts over rolling 400-day windows and horizons of 1--14 days. Log-RMSE gain is combined with forecast-error distribution entropy (FEDE), which captures horizon-dependent changes in the signed error distribution. The resulting \emph{directional predictive signature} comprises response class, RMSE effect, FEDE endpoint and early-slope responses, persistence across rolling regimes, and horizon profile. For screening candidate directions, we also introduce a relative persistence-weighted directional predictive response strength (DPRS) on a within-study 0--100 scale; 100 denotes the strongest observed response in the comparison set, not an absolute coupling maximum. The pairs show concordant, RMSE-only, FEDE-only, and mixed responses. ``loss of smell''$\rightarrow$reported cases is accuracy-dominant, with mean log-RMSE gain 0.104 (about 9.9\% equivalent RMSE reduction) and DPRS 100, whereas the reverse direction is FEDE-dominant ($D_{\FEDE}=0.048$; DPRS $=42$). ``covid test''$\rightarrow$positive rate is FEDE-dominant ($D_{\FEDE}=0.039$; positive in 72\% of windows; DPRS $=98$) despite negative RMSE gain, while ``lockdown''$\rightarrow$stringency is the clearest concordant case (DPRS $=78$). World-level symptom/case pairs are less persistent. Thus, forecast accuracy and error-distribution dynamics are complementary, predictive effects can be direction- and metric-dependent, and forecast horizon is part of the relationship. Aggregated regime-level patterns should not be extrapolated to individual time points or interpreted as evidence of causality.
\end{abstract}

\medskip

\normalsize
\section{Introduction}\label{sec:introduction}

\normalsize
\subsection{Web search activity as an information signal during epidemics}

The COVID-19 pandemic created an unusually rich setting in which epidemiological processes, public information seeking, media exposure, policy decisions, and individual behavior evolved simultaneously. Search engines record one observable layer of this coupled system. A query about symptoms, testing, vaccination, or restrictions is not a clinical event in itself, but it is a time-stamped trace of information demand generated by people who may be symptomatic, concerned, exposed to news, planning an action, or reacting to an already visible epidemic wave. This dual character makes web-search data attractive for digital epidemiology and information science: search activity can contain early information about later epidemiological observations, while epidemiological events and public communication can themselves stimulate subsequent search behavior.

A large literature has therefore investigated whether Google Trends and related search data track or anticipate COVID-19 incidence, symptoms, hospitalization burden, and other measures. Systematic and narrative reviews conclude that search data often provide useful supplementary signals, but also emphasize substantial heterogeneity in query choice, geography, pandemic phase, preprocessing, evaluation design, and forecasting horizon \citep{Ma2022Review,Saegner2022,Rovetta2021}. Early studies reported associations between symptom-related searches and subsequent case counts in the United States, Europe, India, Africa, and multi-country settings \citep{Mavragani2020,Venkatesh2020,Fulk2021,Lu2021}. Later work used machine learning and time-varying models to improve case prediction or to adapt search features as public interest changed \citep{Rabiolo2021,Habibdoust2024,Ahn2024}. These findings established online information seeking as a potentially useful component of epidemic surveillance, but they did not establish a single stable mapping from search volume to disease burden.

The distinction between \emph{association} and \emph{incremental predictive information} is crucial. Two series can be strongly correlated because they share a trend, a seasonal component, an external driver, or a common information environment. A stronger operational question is whether the past of a candidate source improves prediction of a target beyond the information already contained in the target's own history. This is the logic underlying Granger-type predictive analysis \citep{Shojaie2021}. In forecasting terms, it motivates a matched comparison between a restricted model using only target history and an augmented model that also receives the candidate source. Out-of-sample loss then provides a direct measure of whether the auxiliary series adds predictive value, while formal forecast-comparison procedures address the statistical evidence for differences in forecast loss \citep{Diebold1995}. The present study adopts this incremental-prediction perspective but extends it in two ways: the comparison is explicitly bidirectional, and predictive response is characterized across multiple forecast horizons rather than collapsed into a single error statistic.

\normalsize
\subsection{What is already known: early-warning value, temporal heterogeneity, and reverse effects}

Among the most relevant findings for the present work is evidence that symptom-related searches can precede later hospital burden. In the United States, internet-search information added to autoregressive epidemic signals improved retrospective out-of-sample forecasts of new hospital admissions at one- and two-week horizons; importantly, the search contribution became weaker as the forecast horizon increased \citep{Wang2022Hospital}. In England, analyses across NHS Trusts found Google Trends and other digital indicators to lead hospital admissions in many locations, with typical leads of roughly 5--20 days, but the magnitude and even usefulness of these leads changed across waves and regions \citep{Mellor2023Leading}. Other hospital-oriented studies likewise support the value of digital leading indicators while showing that gains are context dependent rather than universal \citep{Mellor2023Realtime,Wang2023Alert}. Thus, the literature supports the interpretation of symptom searches as supplementary short-term leading signals, not as stable or causal precursors of hospitalization.

The loss-of-smell literature provides an especially instructive example because both temporal directions are plausible. Multiple studies reported that searches for anosmia, loss of smell, or loss of taste preceded reported cases by approximately one to three weeks in at least some periods and countries \citep{Chen2022,Lippi2020Smell,Walker2020,Cherry2020}. Forecasting studies found that such terms could improve short-horizon case prediction, although their performance was not uniformly superior to broader respiratory symptoms \citep{Amusa2022,Porcu2023}. At the same time, other analyses found smell-related searches to coincide with or follow case increases, especially after major media attention made anosmia widely recognized as a COVID-19 symptom \citep{Higgins2020,Wu2020,Panuganti2020,SousaPinto2020,Madden2021}. These observations are not necessarily contradictory. They are consistent with two distinct observational pathways. In a symptom-led pathway, infection may be followed by anosmia, symptom-related information seeking, subsequent testing or diagnosis, and only then appearance in reported case counts. In an attention-led pathway, rising reported cases may increase news coverage and public attention, raise awareness of anosmia, and thereby stimulate smell-related searches among a broader population. The two pathways can therefore produce opposite temporal ordering without requiring one universal direction of the relationship.

This example exposes a broader limitation of interpreting a single lead, correlation coefficient, or forecast score as ``the'' direction of a relationship. Search behavior is generated by a mixture of symptom experience, risk perception, news exposure, policy announcements, and intended actions. Correspondingly, the same pair of observed time series may contain predictive information in both directions, and the two directions need not be expressed in the same way. A source may lower the magnitude of future forecast errors in one direction, while the reverse source may mainly change how forecast uncertainty develops with increasing horizon. The literature reviewed for this study contains many forward lead--lag analyses but comparatively few designs that estimate the two directions symmetrically under the same predictor, data window, lag history, and evaluation protocol. This gap is particularly explicit in hospitalization studies, where reverse hospitalization-to-search tests are rarely conducted within the same modeling design \citep{Mellor2023Leading,Wang2022Hospital,Sato2021}.

Forecast horizon is another unresolved dimension. Published studies repeatedly show that useful search information is concentrated at short or intermediate horizons and can deteriorate as the horizon expands \citep{Wang2022Hospital,Chen2022,Amusa2022,Yang2023,Radziwon2023}. However, horizon dependence is commonly summarized by reporting which lag or forecasting horizon performed best. That practice can miss a different phenomenon: adding a source series can alter the \emph{shape} of the forecast-error distribution across horizons even when the mean forecast error does not improve. Such a response is potentially important in observational systems where predictive information is distributed across heterogeneous behavioral delays rather than concentrated in one deterministic lag.

Finally, the Google Trends literature contains strong methodological warnings. More rigorous time-series treatment has eliminated a substantial fraction of apparently predictive symptom terms found by simple correlation, demonstrating the risks of nonstationarity and multiple testing \citep{Sato2021}. Media coverage can generate sharp search-volume changes unrelated to local infection dynamics, particularly for highly publicized symptoms such as loss of smell and taste \citep{SousaPinto2020,Panuganti2020,Higgins2020}. Search--disease relationships also vary across epidemic phases and locations as testing availability, public knowledge, internet access, and health-seeking behavior change \citep{Asseo2020,Husnayain2021,Fulk2021,Asplin2025,Marty2023}. These limitations argue against causal language based on observational search data and motivate a model-relative interpretation in terms of \emph{directional predictive contribution}.

Table~\ref{tab:literature_synthesis} summarizes the literature landscape that motivates the present analysis. The key point is not that prior findings are inconsistent, but that several different phenomena have often been projected onto one scalar notion of ``predictive value.''

\par\medskip
\refstepcounter{table}\label{tab:literature_synthesis}
\noindent\textbf{Table \thetable}\\[-1mm]
\noindent\small Synthesis of the literature most directly motivating the present study. The table separates what is reasonably established from what remains unresolved for a symmetric, multi-horizon predictive analysis.\par\smallskip
\scriptsize
\begin{tabularx}{\textwidth}{@{}p{3.0cm}Y Y Y@{}}
\toprule
Topic & What is reasonably established & Main limitation of prior evidence & Question retained here \\
\midrule
Symptom searches and cases & Symptom-related queries often track or precede reported cases across countries and can improve selected forecasting models \citep{Lampos2020,Lu2021,Venkatesh2020,Mavragani2020,Habibdoust2024}. & Lead times and useful keywords drift across geography, waves, and information environments. & Does auxiliary search history add persistent information beyond the target's own history under one common forecasting protocol? \\
\addlinespace
Searches and hospitalization & Google/search indicators can improve short-term hospitalization forecasts; reported leads are commonly of order days to a few weeks \citep{Wang2022Hospital,Mellor2023Leading,Mellor2023Realtime,Wang2023Alert}. & Direct hospitalization studies are fewer than case studies; reverse direction is rarely tested symmetrically. & Is search-to-hospitalization stronger than hospitalization-to-search when both directions use exactly the same design? \\
\addlinespace
Loss of smell and cases & Anosmia-related searches often precede case counts by roughly 1--3 weeks and sometimes improve forecasts \citep{Chen2022,Lippi2020Smell,Walker2020,Amusa2022}. & Strong media and phase effects can reverse or erase the apparent lead. & Can opposite directions be separated by their different effects on forecast accuracy and error-distribution dynamics? \\
\addlinespace
Reverse information flow & Case surges, media reports, and public awareness can stimulate subsequent COVID-related search activity \citep{Higgins2020,SousaPinto2020,Wu2020,Madden2021}. & Forward and reverse mechanisms are often demonstrated in different studies, periods, and models. & Can both directions be evaluated within one matched source--target framework? \\
\addlinespace
Forecast-horizon dependence & Search information is often most useful at short/intermediate horizons and weakens at longer horizons \citep{Wang2022Hospital,Chen2022,Yang2023,Radziwon2023}. & A ``best lag'' or aggregate RMSE does not describe how the full error distribution changes with horizon. & What additional information is revealed by horizon-resolved forecast-error entropy? \\
\addlinespace
Media, awareness, and nonstationarity & Media exposure, public knowledge, testing regimes, and changing behavior can produce false or unstable associations \citep{Sato2021,SousaPinto2020,Asseo2020,Husnayain2021,Asplin2025}. & These processes make causal interpretation unsafe and can generate transient peaks. & Can long rolling windows identify persistent predictive regimes without interpreting individual dates as causal events? \\
\bottomrule
\end{tabularx}\par\medskip
\normalsize

\normalsize
\subsection{From forecast accuracy to forecast-error distribution dynamics}

Conventional forecast evaluation asks whether access to a candidate source makes the target forecast more accurate. In the present study, this is quantified by a mean multi-horizon log-RMSE gain: positive values indicate that the augmented model has lower RMSE than the target-only baseline. This quantity is intuitive and operationally important, but it reduces each horizon to one second-moment summary. Two models may have similar RMSE while producing errors with different signs, dispersion patterns, concentration, or horizon evolution.

The methodological basis of the present analysis is the recently introduced forecast-error distribution entropy (FEDE) framework \citep{VelichkoFEDE2026}. FEDE was developed to assess multi-horizon nonlinear predictive coupling by comparing matched target-only and target-plus-source forecasts and by characterizing the complete signed forecast-error distribution at each horizon. The corresponding reusable numerical implementation and controlled synthetic benchmark suite are archived as a citable software release \citep{VelichkoFEDESoftware2026}. Its controlled validation included known source--target coupling, nonlinear transformations, additive noise, reverse-direction and independent-source controls, alternative model forms, and broad variations of window length, history length, and maximum forecast horizon. In those controlled systems, conventional forecast gain and FEDE responses tended to move in a common direction because a single imposed interaction channel drove all descriptors. The synthetic study therefore established sensitivity and robustness of the metric; the present study asks a different question, namely whether FEDE remains interpretable when the data are observational and predictive dependence may combine behavioral response, epidemiological dynamics, policy, media attention, common forcing, and feedback.

Accordingly, the synthetic validation is not repeated here. We use FEDE as an established methodological instrument and focus on the structure of bidirectional predictive responses in Google Trends and COVID-19 data. Observational data need not preserve the one-dimensional calibration picture of the benchmark systems. Under multiple delays and feedback pathways, improvement in average error magnitude and reorganization of the multi-horizon error distribution are related but non-equivalent properties.

This distinction suggests a concrete interpretation of metric disagreement. If both RMSE gain and FEDE responses are persistently positive in one direction and weak in reverse, the evidence supports a relatively coherent directional predictive contribution. If RMSE improves but FEDE changes little, the auxiliary series may sharpen point prediction without strongly changing the geometry of errors across horizons. Conversely, a pronounced FEDE response with little or negative RMSE gain indicates that the auxiliary history changes the horizon-dependent structure of prediction errors without improving average accuracy. Finally, opposite directions of the same pair may be expressed through different metrics. We refer to this latter pattern as \emph{metric-dependent directional asymmetry}: bidirectional predictive coupling is present, but the two directions are not equivalent in how they alter future prediction.

This perspective is particularly relevant to information behavior. A search query can be an early trace of an intended action, a symptom-driven reaction, or a response to already observed events. These mechanisms can operate simultaneously and on different timescales. Therefore, the scientifically relevant question is not simply whether ``Google predicts COVID,'' but whether the addition of search history changes the predictability of an epidemiological variable, how that change evolves from short to longer forecast horizons, and whether the reverse direction exhibits the same or a different response.

\normalsize
\subsection{Present study and contributions}

We study eight pre-specified pairs linking Google search activity with epidemiological or policy indicators related to COVID-19. They were selected to represent distinct, literature-motivated forms of information--epidemic interaction: symptom searches and hospitalization, testing interest and test positivity, loss-of-smell searches and reported cases, vaccine-related searches and vaccination activity, corresponding world-level symptom/case pairs, lockdown searches and policy stringency, and a flu-search/COVID-case pair used as a seasonal or nonspecific stress case. The pairs were fixed before the final $W=400$ analysis rather than selected from the final results. Each pair is evaluated in both directions---search $\rightarrow$ paired indicator and paired indicator $\rightarrow$ search---using the same LightGBM model class, feature history, forecast horizons, chronological split, and rolling-window protocol. This symmetry is central to the design: differences between directions should reflect differences in incremental predictive information rather than differences in model or validation procedure.

The analysis uses 400-day rolling windows, a 14-day history, forecast horizons $h=1,\ldots,14$, a one-day rolling step, and a chronological 60/40 train--test split. The long window is not used to localize a predictive effect to a particular calendar day. Adjacent windows share almost all observations, so individual rolling points are strongly dependent. Instead, the design targets broad regimes in which the relationship between the paired series persists sufficiently long to produce a stable multi-horizon response. Temporal fractions of positive windows are therefore interpreted as persistence or coverage measures rather than as counts of independent replications. For the principal cross-window summaries, the analysis averages horizon-specific responses over the available rolling windows and reports interquartile ranges to retain information about temporal heterogeneity.

The study makes five linked contributions. First, it distinguishes \emph{forecast-accuracy gain} from \emph{forecast-error distribution dynamics}. Mean log-RMSE gain measures whether the auxiliary history reduces average prediction error, whereas FEDE measures whether it reorganizes the signed error distribution across forecast horizons. The two responses are therefore treated as complementary rather than interchangeable. Second, it places forward and reverse directions in one matched framework and tests whether the two directions are expressed through the same forecast property. This allows identification of \emph{metric-dependent bidirectional asymmetry}, in which one direction is dominated by point-accuracy gain and the opposite direction by FEDE response. Third, it treats forecast horizon as part of the predictive relation itself: the complete $h=1,\ldots,14$ profiles are retained so that increasing, decreasing, intermediate-horizon, and otherwise nonuniform responses are not hidden by one aggregate score or one selected lag. Fourth, it interprets the resulting effects at the regime level. Aggregation across broad $W=400$ windows is used to identify dominant multi-horizon profiles, while temporal context plots and interquartile ranges explicitly show that local behavior can vary around those dominant patterns. For interpretation, a first-level sign map later separates concordant, RMSE-only, FEDE-only, and no-positive-response patterns; directions with internally inconsistent FEDE summaries are retained as mixed rather than forced into a class. These labels describe response type, not effect size. Fifth, it turns the multicomponent analysis into a practical screening output. Each ordered pair is summarized by a \emph{directional predictive signature} containing response class, equivalent RMSE effect, FEDE endpoint and early-slope effects, persistence across rolling regimes, and the dominant horizon profile. A secondary relative DPRS score combines positive response magnitude with sign persistence to rank candidate directions within the same comparison set. The score is deliberately comparative: it does not replace the signature, determine class membership, or define an absolute scale of coupling strength.

The practical objective is therefore not to reduce each pair to a single coupling coefficient, but to construct a directional predictive signature that preserves response type, magnitude, persistence, and horizon dependence; relative DPRS adds a secondary within-set ordering when one number is useful for screening.

These contributions lead to four defensible propositions that organize the Results and Discussion. (i) Forecast-accuracy gain and FEDE can encode different but complementary forms of incremental predictive information. (ii) Bidirectional predictive coupling need not be symmetric in the metric through which it is expressed. (iii) The sign and magnitude of incremental predictive information can depend on forecast horizon. (iv) Aggregated regime-level patterns should not be extrapolated to individual time points or interpreted as evidence of causality. This framing deliberately avoids the stronger claim that search behavior physically causes the corresponding epidemiological or policy variable.

Table~\ref{tab:design_gap} makes the relationship between recurring literature gaps and the present design explicit. The remainder of the paper is organized as follows. Section~\ref{sec:data} describes construction of the Google Trends and OWID series and the eight pre-specified pairs. Section~\ref{sec:methods} defines the matched bidirectional forecasting design, RMSE gain, FEDE responses, and aggregation across rolling windows. Section~\ref{sec:results} presents temporal examples, cross-pair summaries, dominant horizon profiles, and representative error distributions. Section~\ref{sec:discussion} interprets the observed metric agreement and disagreement in relation to the synthetic FEDE calibration and the Google Trends/COVID-19 literature, and Section~\ref{sec:limitations} states the main limitations before the conclusions in Section~\ref{sec:conclusion}.

\par\medskip
\refstepcounter{table}\label{tab:design_gap}
\noindent\textbf{Table \thetable}\\[-1mm]
\noindent\small How the present design addresses recurring limitations in the Google Trends/COVID literature.\par\smallskip
\scriptsize
\begin{tabularx}{\textwidth}{@{}p{3.6cm}Y Y@{}}
\toprule
Recurring limitation & Present design choice & Intended interpretation \\
\midrule
Correlation or lag alone does not establish incremental predictive value & Matched target-only and target-plus-source LightGBM forecasts & Measures information added by the source beyond target history \\
\addlinespace
Direction often studied only as search $\rightarrow$ epidemiological outcome & Every pair is recomputed in the reverse direction & Separates directional asymmetry from simple association \\
\addlinespace
One aggregate forecast score can hide horizon-dependent structure & Log-RMSE gain is complemented by FEDE change and early-slope response over $h=1,\ldots,14$ & Distinguishes average error reduction from redistribution of errors across horizons \\
\addlinespace
A selected ``best lag'' can oversimplify heterogeneous behavioral delays & Full multi-horizon profiles are retained & Identifies where predictive contribution strengthens, weakens, or changes character \\
\addlinespace
Short local peaks are sensitive to wave timing, media events, and noise & Rolling windows use $W=400$ days with points displayed at window centers & Emphasizes persistent regimes; does not assign a causal event to a single plotted date \\
\addlinespace
Google Trends is vulnerable to media, awareness, seasonality, and common forcing & Reverse-direction analysis, a nonspecific/seasonal stress-test pair, and regime-level interpretation & Supports cautious interpretation as predictive coupling rather than causation \\
\addlinespace
Multicomponent responses can be difficult to prioritize across many candidate series & Directional predictive signature plus relative persistence-weighted DPRS ranking & Supports within-study screening while preserving response type, persistence, and horizon structure \\
\bottomrule
\end{tabularx}\par\medskip
\normalsize

The central hypothesis is therefore not that all three metrics should agree. Agreement is expected in relatively simple or strongly directional cases, whereas disagreement can itself be informative in observational data. The empirical question is whether recurring patterns of agreement and disagreement correspond to plausible differences between direct forecasting benefit, delayed behavioral response, feedback-like information flow, and nonspecific shared dynamics. This framing turns metric divergence from a nuisance into an object of study and provides the organizing principle for the remainder of the paper.

\normalsize
\section{Data and study design}\label{sec:data}

The empirical analysis has two distinct stages that should not be conflated. First, an acquisition and harmonization pipeline constructs continuous daily Google Trends and OWID series on compatible calendar dates. Second, the main FEDE experiment takes the resulting paired daily series as its input and performs rolling bidirectional forecasting. The first stage concerns data construction only; no FEDE or forecast-error statistic is calculated during Google Trends stitching or OWID preparation. Appendix~\ref{app:data_pipeline} and Fig.~\ref{fig:data_pipeline} summarize this acquisition pipeline, whereas the forecasting workflow used after the series have been constructed is described in Section~\ref{sec:methods} and Fig.~\ref{fig:main_workflow}.

\normalsize
\subsection{Google Trends data acquisition and harmonization}\label{sec:google_data}

To create the search series used as inputs to the forecasting experiment, daily Google Trends data were assembled for 1 September 2019 through 1 December 2024 for two geographical scopes: the United States and the worldwide Google Trends scope. Google Web Search was used. Ten queries were acquired in two five-query groups: ``covid'', ``coronavirus'', ``covid symptoms'', ``covid test'', ``covid vaccine'', ``loss of smell'', ``lockdown'', ``quarantine'', ``face mask'', and ``flu''. The final analysis uses a subset of these queries defined by the search--indicator pairs in Table~\ref{tab:pairs}.

Google Trends does not provide absolute search counts. It returns a relative search-interest index whose scale is normalized within the particular requested time interval. Consequently, values obtained from independently downloaded intervals are not automatically on a common scale, and a long multi-year series cannot be reconstructed by simply concatenating separate requests. In addition, sufficiently long requests may be returned at weekly rather than daily resolution. To construct a consistent daily series over the complete study interval, the acquisition pipeline therefore divided the period into overlapping 250-day chunks with 45-day overlap and rejected a chunk if its returned temporal resolution was not approximately daily. The common 45-day segments serve only to estimate how adjacent Google Trends chunks should be rescaled relative to one another; they are unrelated to the later $W=400$ FEDE rolling windows. Adjacent chunks were then placed on a common scale from their shared overlap. For each query with positive values in both chunks, a robust scale factor was estimated from the median log-ratio after trimming the 10th and 90th percentiles of the overlap distribution when sufficient observations were available,
\begin{equation}
 s=\exp\!\left[\operatorname{median}\!\left(\log\frac{x_{\mathrm{old}}}{x_{\mathrm{new}}}\right)\right].
 \label{eq:gt_scale}
\end{equation}
The newer chunk was multiplied by $s$, overlapping observations were averaged, and the procedure was repeated sequentially. After stitching, each query $q$ was normalized independently over the full assembled period,
\begin{equation}
 I_q(t)=100\,\frac{X_q(t)}{\max_t X_q(t)}.
 \label{eq:gt_norm}
\end{equation}
Thus, the 0--100 values describe relative temporal interest within a query. They are not absolute search counts and should not be interpreted as literal amplitude comparisons between different queries. The frozen USA and world Google Trends tables each contain 1919 daily observations. At this point the acquisition stage is complete: the subsequent FEDE analysis uses these merged daily series as ordinary observed time series and does not operate on the original 250-day download chunks.

\normalsize
\subsection{COVID-19 and policy data}\label{sec:owid_data}

Epidemiological and policy variables were derived from a frozen Our World in Data (OWID) COVID-19 archive dated 19 August 2024. Freezing the source archive was important because historical OWID values can be retrospectively revised; the article calculations therefore use fixed compact input tables rather than a later redownload. The resulting USA table contains daily observations from 5 January 2020 through 4 August 2024, and the global candidate table spans 1 January 2020 through 14 August 2024.

For additive global quantities such as cases, deaths, and vaccination counts, the official OWID World series was preferred whenever available. An independent country-level aggregation was retained as a validation route. For selected non-additive indicators, including reproduction rate, test positivity, tests per case, and policy stringency, global values were constructed as population-weighted country means. Cumulative country variables were forward-filled only after their first valid report, whereas daily flows, hospital/ICU stocks, and admission variables were not forward-filled. Coverage fractions were retained during global aggregation so that sparse reporting was not hidden.

\normalsize
\subsection{Series representation, alignment, and leakage control}\label{sec:representation}

The forecasting models receive the prepared daily \emph{levels} of the Google Trends and OWID variables. No additional log transformation, log return, percentage change, or first difference is applied to the input series. Some OWID variables used as targets, such as \texttt{new\_cases\_smoothed} and \texttt{new\_vaccinations\_smoothed}, are already smoothed quantities supplied by OWID; this is part of the source-variable definition rather than an additional smoothing step introduced by the present method.

For each search--indicator pair, observations are aligned by calendar date, and only complete daily windows are admitted to the rolling experiment. Missing dates are not compressed into artificial consecutive observations. Within each rolling window, feature columns are standardized using statistics estimated from training forecast origins only. For feature $j$,
\begin{equation}
 z_j=\frac{x_j-\mu_j}{\sigma_j},
 \label{eq:standardize}
\end{equation}
where $\mu_j$ and $\sigma_j$ are computed from the training origins of that window, with a numerical standard-deviation floor of $10^{-12}$. The same training estimates are applied to held-out features; test observations do not contribute to scaling. The prediction target remains on its original numerical scale.

\normalsize
\subsection{Pre-specified search--indicator pairs}\label{sec:pairs}

Eight representative pairs were fixed before the final $W=400$ analysis and carried forward from an earlier exploratory Google/COVID benchmark. They were chosen to span clinical symptom signals, testing behavior, vaccination behavior, policy/information feedback, world-level generalization checks, and a seasonal/nonspecific stress case. They were not selected from the final $W=400$ results to maximize effect size. Every pair is evaluated in both directions under the same forecasting protocol.

\par\medskip
\refstepcounter{table}\label{tab:pairs}
\noindent\textbf{Table \thetable}\\[-1mm]
\noindent\small Pre-specified Google-search/indicator pairs, actual continuous paired intervals, and numbers of complete $W=400$ rolling windows used in the bidirectional analysis. ``Forward'' denotes search $\rightarrow$ paired indicator; ``reverse'' denotes paired indicator $\rightarrow$ search. The two series are exchanged without changing the forecasting protocol.\par\smallskip
\scriptsize
\begin{tabularx}{\textwidth}{@{}p{0.5cm}p{0.75cm}p{1.8cm}p{2.7cm}p{2.45cm}p{0.65cm}p{0.75cm}Y@{}}
\toprule
ID & Scope & Search series & Epidemiological/policy series & Continuous paired interval & Days & Windows & Role in the study \\
\midrule
B1 & USA & ``covid symptoms'' & hospital patients (\texttt{hosp\_patients}) & 15 Jul 2020--25 Apr 2024 & 1381 & 982 & Clinical symptom information $\leftrightarrow$ hospitalization burden \\
B2 & USA & ``covid test'' & test positivity (\texttt{positive\_rate}) & 7 Mar 2020--18 Jun 2022 & 834 & 435 & Testing information demand $\leftrightarrow$ positivity \\
B3 & USA & ``loss of smell'' & smoothed new cases & 10 Jan 2020--20 May 2023 & 1227 & 828 & Symptom information $\leftrightarrow$ reported incidence \\
B4 & USA & ``covid vaccine'' & smoothed new vaccinations & 14 Dec 2020--9 May 2023 & 877 & 478 & Vaccine information demand $\leftrightarrow$ vaccination activity \\
B5 & World & ``covid symptoms'' & smoothed new cases per million (OWID World) & 10 Jan 2020--4 Aug 2024 & 1669 & 1270 & Global symptom-search/generalization check \\
B6 & World & ``loss of smell'' & smoothed new cases per million (OWID World) & 10 Jan 2020--4 Aug 2024 & 1669 & 1270 & Global anosmia-search/generalization check \\
B7 & USA & ``lockdown'' & stringency index & 5 Jan 2020--31 Dec 2022 & 1092 & 693 & Public information demand $\leftrightarrow$ policy stringency \\
B8 & USA & ``flu'' & smoothed new cases & 10 Jan 2020--20 May 2023 & 1227 & 828 & Seasonal/nonspecific stress-test pair \\
\bottomrule
\end{tabularx}\par\medskip
\normalsize

The pair-specific intervals in Table~\ref{tab:pairs} are the longest continuous calendar-day spans on which both selected variables are simultaneously available after date alignment. Their different endpoints therefore arise from variable-specific OWID availability rather than from truncating the Google Trends series. For a complete paired interval of $N$ days, the one-day rolling design yields $N-W+1$ complete $W=400$ windows, which gives the counts reported in the table.

\normalsize
\section{Methods}\label{sec:methods}

Once the two daily series for a pair have been constructed and aligned, the data-acquisition machinery no longer enters the calculation. The main experiment operates only on a source series $Y$ and a target series $X$. Figure~\ref{fig:main_workflow} summarizes the complete bidirectional analysis: the same rolling window and horizon structure is used to compare a target-only forecast with a target-plus-source forecast, the resulting held-out signed errors are evaluated by both RMSE-based and FEDE-based descriptors, and the complete calculation is then repeated after exchanging the two series. This matched design makes the comparison between forward and reverse directions directly interpretable within one model class and one validation protocol.

\par\medskip
\refstepcounter{figure}\label{fig:main_workflow}
\begin{center}
\includegraphics[width=0.98\textwidth]{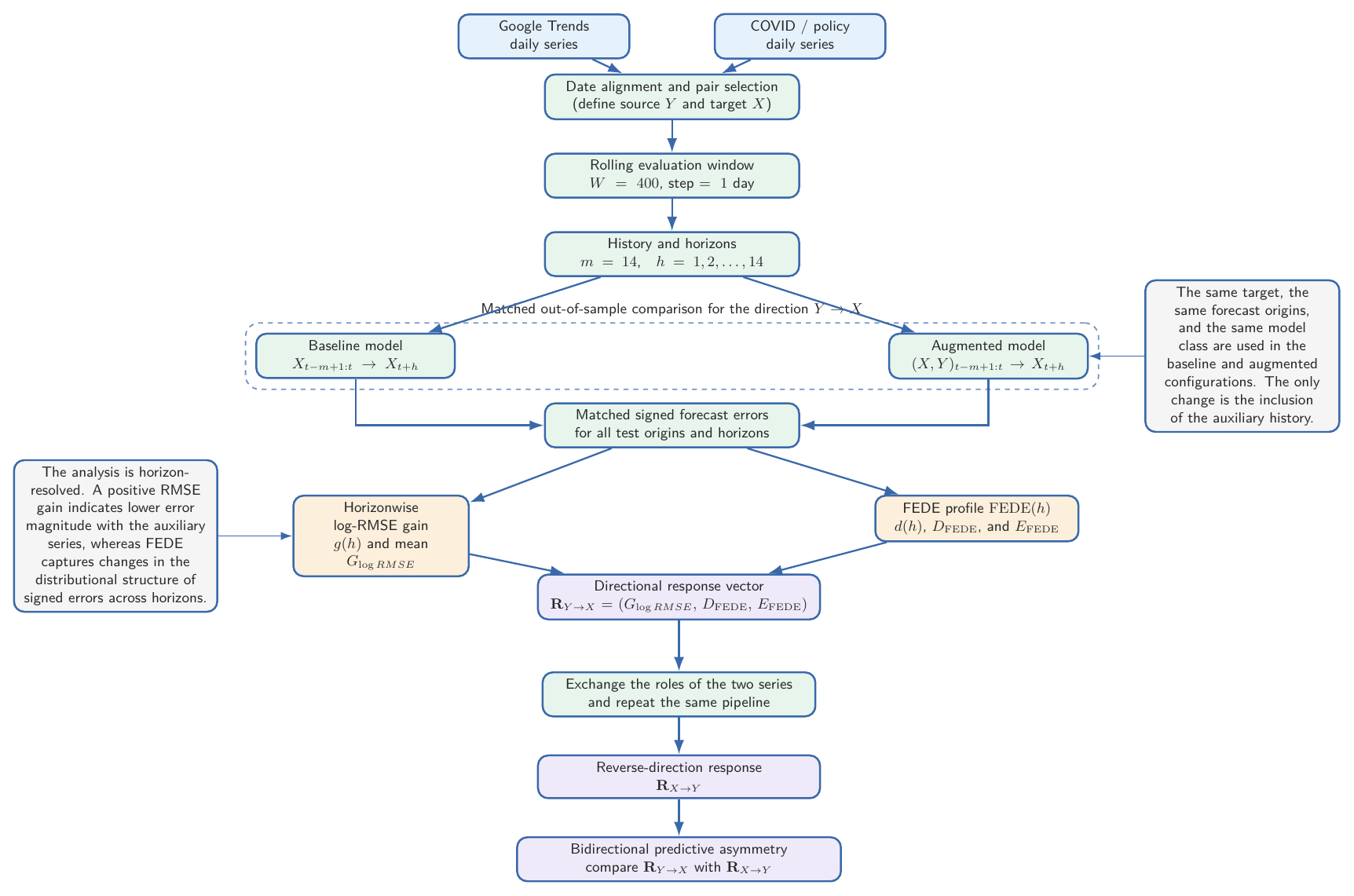}
\end{center}
\noindent\small\textbf{Figure \thefigure.} Main bidirectional multi-horizon forecasting workflow. The inputs are already prepared and calendar-aligned daily Google Trends and COVID/policy series. For each direction, the baseline and augmented LightGBM branches share the same target, forecast origins, rolling window, history, horizons, and chronological train/test split; they differ only in whether the candidate-source history is included. Horizon-resolved RMSE gain and FEDE responses form the directional response vector, after which source and target are exchanged and the same procedure is repeated to characterize bidirectional predictive asymmetry.\par\medskip
\normalsize

\normalsize
\subsection{Predictive question and matched forecasting branches}\label{sec:matched}

Let $x_t$ denote the target series and $y_t$ a candidate source. The experiment asks whether the recent history of $y$ contributes out-of-sample information about future $x$ beyond the information already available in the recent history of $x$. This is a model-relative predictive question in the Granger sense and does not by itself establish causality.

For a history length $m$, define
\begin{equation}
 \mathbf{x}_t=(x_{t-m+1},\ldots,x_t), \qquad
 \mathbf{y}_t=(y_{t-m+1},\ldots,y_t).
\end{equation}
For each forecast horizon $h=1,\ldots,H$, two separate regression branches predict the identical future target $x_{t+h}$. The restricted or baseline branch uses only target history,
\begin{equation}
 \widehat{x}^{(b)}_{t+h}=f_h(\mathbf{x}_t),
 \label{eq:baseline}
\end{equation}
whereas the augmented branch adds the candidate-source history,
\begin{equation}
 \widehat{x}^{(a)}_{t+h}=g_h(\mathbf{x}_t,\mathbf{y}_t).
 \label{eq:augmented}
\end{equation}
The two branches use exactly the same forecast origins, training/test partition, target observations, and prediction engine. Thus, their difference isolates the incremental contribution of source history within the specified model class. The reverse direction is obtained by exchanging $x$ and $y$ and repeating the complete procedure. In the forward ``search $\rightarrow$ COVID'' direction this means comparing $x\rightarrow x$ with $(x,y)\rightarrow x$; in reverse it means comparing $y\rightarrow y$ with $(y,x)\rightarrow y$.

\normalsize
\subsection{Rolling windows and chronological validation}\label{sec:rolling}

The final main analysis uses rolling windows of $W=400$ calendar days, target/source history $m=14$ days, maximum forecast horizon $H=14$ days, and a one-day rolling step. A complete window contains
\begin{equation}
 N_{\mathrm{orig}}=W-m-H+1=373
\end{equation}
legal forecast origins that are common to all horizons. Keeping the origin set fixed prevents apparent changes with $h$ from being created by changing training or test samples. Origins are split chronologically: the first 60\% (223 origins) are used for training and the remaining 40\% (150 origins) for held-out evaluation. No random shuffle is used.

A one-day rolling step denotes one actual calendar day. Windows containing gaps in the selected source or target are not converted into shorter pseudo-consecutive sequences. Because adjacent complete windows share 399 of 400 calendar days, neighboring rolling estimates are strongly dependent. Percentages of positive windows are therefore interpreted as persistence or temporal coverage descriptors, not as counts of independent replications or as inputs to a binomial significance argument.

\normalsize
\subsection{Prediction engine and fixed model configuration}\label{sec:lgbm}

Both forecasting branches use LightGBM gradient-boosted regression trees \citep{Ke2017}. Hyperparameters are fixed for all pairs, both directions, all rolling windows, and all horizons; they are not re-optimized to maximize a pair-specific effect. The fixed configuration is summarized in Table~\ref{tab:lgbm_settings}. Independent pair-direction calculations are parallelized externally, while each LightGBM fit uses one internal thread.

\par\medskip
\refstepcounter{table}\label{tab:lgbm_settings}
\noindent\textbf{Table \thetable}\\[-1mm]
\noindent\small Final LightGBM and FEDE settings used throughout the main analysis.\par\smallskip
\scriptsize
\begin{tabularx}{\textwidth}{@{}Y p{2.8cm}Y p{2.8cm}@{}}
\toprule
Parameter & Value & Parameter & Value \\
\midrule
Number of trees & 50 & Number of leaves & 7 \\
Maximum depth & 3 & Learning rate & 0.05 \\
Minimum child samples & 5 & $L_1$ regularization & 0 \\
$L_2$ regularization & 0.1 & Entropy bins & 16 \\
History length $m$ & 14 days & Maximum horizon $H$ & 14 days \\
Rolling window $W$ & 400 days & Rolling step & 1 day \\
Train/test split & 60/40 chronological & Numerical floors & $10^{-12}$ \\
\bottomrule
\end{tabularx}\par\medskip
\normalsize

\normalsize
\subsection{Horizon-specific and mean log-RMSE gain}\label{sec:rmse}

Let $e_b(i,h)=x_{i+h}-\widehat{x}^{(b)}_{i+h}$ and $e_a(i,h)=x_{i+h}-\widehat{x}^{(a)}_{i+h}$ denote held-out signed errors for the baseline and augmented branches at horizon $h$. Their horizon-specific root-mean-square errors are
\begin{equation}
 \RMSE_b(h)=\sqrt{\frac{1}{N}\sum_i e_b(i,h)^2}, \qquad
 \RMSE_a(h)=\sqrt{\frac{1}{N}\sum_i e_a(i,h)^2}.
 \label{eq:rmse}
\end{equation}
Rather than retaining only a single aggregate score, we first define the horizon-resolved log-RMSE gain
\begin{equation}
 g(h)=\log\frac{\RMSE_b(h)+\varepsilon}{\RMSE_a(h)+\varepsilon},
 \label{eq:gh}
\end{equation}
where $\varepsilon=10^{-12}$. Positive $g(h)$ indicates that source history reduces held-out RMSE at horizon $h$ and is therefore interpreted as useful additional predictive information from the source. A value $g(h)=0$ indicates no change in predictive accuracy. Negative $g(h)$ means that the augmented branch performs worse than the target-only baseline at that horizon. Such negative values are retained as diagnostic contrasts, but they are not interpreted as ``negative coupling,'' an opposite causal effect, or evidence for the reverse predictive direction; they simply indicate that adding the source does not improve out-of-sample prediction under the specified model, window, and sample. This interpretation follows the convention used in the preceding synthetic FEDE study \citep{VelichkoFEDE2026}. The conventional scalar response used for summaries is
\begin{equation}
 G_{\log\RMSE}=\frac{1}{H}\sum_{h=1}^{H}g(h).
 \label{eq:mean_gain}
\end{equation}
Accordingly, positive $G_{\log\RMSE}$ denotes lower average multi-horizon RMSE after adding the candidate source, whereas non-positive values provide no evidence of a positive accuracy contribution.

\normalsize
\subsection{Forecast-error distribution entropy}\label{sec:fede}

FEDE complements error magnitude by describing the empirical distribution of the signed held-out errors across forecast horizons. The full motivation, synthetic calibration, nonlinear benchmarks, directional controls, model-form comparisons, and parameter-robustness tests are reported in the methodological study \citep{VelichkoFEDE2026}; here we give the definitions required to reproduce the empirical analysis.

Within each rolling window, baseline and augmented signed errors are pooled to define one common range $[e_{\min},e_{\max}]$. The same $B=16$ equal-width bins are then used for both branches and for every horizon in that window. If $p^{(q)}_j(h)$ is the empirical probability of bin $j$ for branch $q\in\{b,a\}$ at horizon $h$, normalized Shannon entropy \citep{Shannon1948} gives
\begin{equation}
 \FEDE_q(h)= -\frac{\sum_{j=1}^{B}p^{(q)}_j(h)\log p^{(q)}_j(h)}{\log B},
 \qquad 0\leq \FEDE_q(h)\leq 1.
 \label{eq:fede}
\end{equation}
A useful horizon-resolved diagnostic is the direct augmented-minus-baseline difference
\begin{equation}
 d(h)=\FEDE_a(h)-\FEDE_b(h),
 \label{eq:dh}
\end{equation}
which shows where the two error-distribution profiles separate as forecast horizon increases.

Two scalar FEDE responses summarize complementary aspects of this profile. First, the endpoint change for each branch is
\begin{equation}
 \Delta\FEDE_q=\FEDE_q(H)-\FEDE_q(1),
\end{equation}
leading to the augmented-minus-baseline endpoint response
\begin{equation}
 D_{\FEDE}=\Delta\FEDE_a-\Delta\FEDE_b.
 \label{eq:d_fede}
\end{equation}
Second, a least-squares slope $s^{1:5}_{\FEDE,q}$ is fitted to $\FEDE_q(h)$ over the early horizons $h=1,\ldots,5$, and the early-slope response is
\begin{equation}
 E_{\FEDE}=s^{1:5}_{\FEDE,a}-s^{1:5}_{\FEDE,b}.
 \label{eq:e_fede}
\end{equation}
The sign convention follows the FEDE methodological study \citep{VelichkoFEDE2026}. At a single horizon, $d(h)>0$ means that the augmented branch has higher normalized signed-error entropy than the baseline, whereas $d(h)<0$ means the opposite. Likewise, positive $D_{\FEDE}$ or $E_{\FEDE}$ means that the augmented branch exhibits a larger endpoint increase or early-horizon slope, respectively, than the baseline; negative values indicate a change in the opposite direction. Crucially, these signs are not accuracy scores. A negative FEDE response does not mean poorer predictive accuracy, ``negative coupling,'' or reverse causality. The FEDE descriptors quantify how adding source history changes the horizon evolution of the signed-error distribution, and therefore need not share the sign of $G_{\log\RMSE}$ in observational data.

\normalsize
\subsection{Bidirectional response representation and metric-dependent asymmetry}\label{sec:bidirectional}

For compact notation, each directed source--target calculation is represented by the three-component response
\begin{equation}
 \mathbf{R}_{y\rightarrow x}=\left(G_{\log\RMSE},D_{\FEDE},E_{\FEDE}\right)_{y\rightarrow x},
 \label{eq:response_vector}
\end{equation}
and the same quantities are recomputed after exchanging the variables,
\begin{equation}
 \mathbf{R}_{x\rightarrow y}=\left(G_{\log\RMSE},D_{\FEDE},E_{\FEDE}\right)_{x\rightarrow y}.
\end{equation}
This notation is descriptive rather than a new composite statistic. The primary scientific interpretation does not collapse the three components into a norm or a single ``causality score,'' because doing so would remove precisely the metric-dependent structure of interest. Bidirectional predictive asymmetry is therefore assessed component by component and, where informative, from the full horizon profiles $g(h)$ and $d(h)$. A secondary scalar introduced below is used only to rank the persistence-weighted magnitude of \emph{positive} responses within the present dataset; it is not used for class assignment, significance testing, or causal interpretation.

Interpretation proceeds at two levels. The first level is a coarse sign-based map that asks whether a directed calculation contains a positive point-accuracy response and whether it contains a positive FEDE response. To make these categories easy to reference throughout the Results and Discussion, the four first-level response classes are numbered I--IV in Table~\ref{tab:response_matrix}. This level is intentionally simple: it separates the concordant pattern observed most often in the controlled synthetic calibration from cases in which the two response families diverge \citep{VelichkoFEDE2026}. It is descriptive rather than a significance test. In particular, a weak or non-positive FEDE entry in the matrix does not imply that the full FEDE profile is featureless; a substantial negative or mixed response is examined at the second level rather than assigned a causal meaning.

\par\medskip
\refstepcounter{table}\label{tab:response_matrix}
\noindent\textbf{Table \thetable}\\[-1mm]
\noindent\small First-level response classes used throughout the Results and Discussion. The upper-left cell is the closest empirical analogue of the concordant synthetic calibration.\\[1mm]
\scriptsize
\begin{tabularx}{\textwidth}{@{}p{3.1cm}YY@{}}
\toprule
 & \textbf{FEDE positive} & \textbf{FEDE weak / non-positive} \\
\midrule
\textbf{RMSE positive} & \textbf{Class I -- Concordant response} (synthetic-like): positive RMSE gain and positive FEDE response & \textbf{Class II -- RMSE-only response}: RMSE improves while FEDE shows no sustained positive response \\
\textbf{RMSE weak / non-positive} & \textbf{Class III -- FEDE-only response}: positive FEDE response without a positive RMSE gain & \textbf{Class IV -- No positive response}: neither summary shows a positive response; inspect the full profiles before further interpretation \\
\bottomrule
\end{tabularx}
\normalsize\par\medskip

Here ``FEDE positive'' is an interpretive shorthand for a sustained positive response in the FEDE summaries and/or the dominant $d(h)$ profile; cases in which the endpoint and early-slope descriptors disagree are not forced into a cell and are resolved from the full horizon profile. The class names are deliberately literal: Class I indicates positive responses in both families, Class II a positive RMSE response without a sustained positive FEDE response, Class III a positive FEDE response without a positive RMSE gain, and Class IV no positive response in either summary. These labels describe the first-level sign pattern only; they do not imply a mechanism or causal interpretation. Class membership is also not an effect-size scale. We impose no universal weak/moderate/strong thresholds because $G_{\log\RMSE}$, $D_{\FEDE}$, and $E_{\FEDE}$ have different numerical scales and because persistence varies across pairs. Response strength is described separately by the magnitude of the mean response, the fraction of positive rolling windows, the interquartile spread, and the full horizon profile. Class IV therefore means only that no positive response is identified by the first-level summaries under the present design; it is not proof that the two observed series are unrelated. The second level of analysis examines $g(h)$, $d(h)$, $D_{\FEDE}$, $E_{\FEDE}$, and their temporal dispersion across rolling windows. It asks whether the response grows, decays, crosses zero, peaks at intermediate horizons, or changes sign. \emph{Metric-dependent bidirectional asymmetry} is then identified when the two directions of the same observed pair map most naturally to different first-level response classes and/or exhibit materially different horizon profiles. Thus, asymmetry means more than unequal response magnitude: opposite directions can be expressed through different forecast-error properties. None of these descriptive classes is, by itself, evidence of causality.

\normalsize
\subsection{Secondary relative persistence-weighted directional response strength}\label{sec:dprs}
The class and the full multi-horizon profile remain the primary outputs, but a practical comparison across many directed pairs benefits from a compact relative ranking quantity. We therefore define a secondary \emph{directional predictive response strength} (DPRS) score for each of the 16 directed calculations. DPRS is explicitly a within-study comparative score: it ranks directions in the current candidate set rather than estimating an absolute physical strength of coupling. It is deliberately restricted to positive responses under the sign convention used above. Negative RMSE gain and negative FEDE responses remain visible in the signature but do not contribute positive strength.

Let $\bar G_i$, $\bar D_i$, and $\bar E_i$ denote the mean rolling responses for directed calculation $i$, and let $p_{iG}$, $p_{iD}$, and $p_{iE}$ be the corresponding fractions of rolling windows with positive response. For these sign counts only, a numerical-zero tolerance $\tau=10^{-12}$ is used: a rolling response $x$ is counted as positive when $x>\tau$, negative when $x<-\tau$, and numerical zero when $|x|\leq\tau$. The raw $G$, $D$, and $E$ values and their means are left unchanged; the tolerance only removes floating-point sign ambiguity and is not an effect-size threshold. To distinguish a persistent directional tendency from approximately balanced sign fluctuations, define the excess-positive persistence transform
\begin{equation}
 \pi(p)=[2p-1]_+,
 \qquad [x]_+=\max(x,0).
 \label{eq:persistence_transform}
\end{equation}
Thus $p=0.5$ contributes no persistent sign preference, whereas $p=1$ gives unit persistence. This is a descriptive transform, not a probability or significance test.

The positive accuracy component is based on the geometric-mean-equivalent relative RMSE reduction associated with the mean log gain,
\begin{equation}
 A_i=
 \frac{[1-\exp(-\bar G_i)]_+\,\pi(p_{iG})}
 {\max_j\{[1-\exp(-\bar G_j)]_+\,\pi(p_{jG})\}}.
 \label{eq:dprs_accuracy}
\end{equation}
The FEDE family contains two complementary summaries, so endpoint and early-slope contributions are normalized separately and then averaged,
\begin{equation}
 F_i=\frac{1}{2}\left[
 \frac{[\bar D_i]_+\,\pi(p_{iD})}{\max_j\{[\bar D_j]_+\,\pi(p_{jD})\}}
 +
 \frac{[\bar E_i]_+\,\pi(p_{iE})}{\max_j\{[\bar E_j]_+\,\pi(p_{jE})\}}
 \right].
 \label{eq:dprs_fede}
\end{equation}
Finally,
\begin{equation}
 S_i=100\,\frac{A_i+F_i}{\max_j(A_j+F_j)}.
 \label{eq:dprs}
\end{equation}
By construction, the strongest persistence-weighted positive response observed in the present comparison set receives $S=100$. Thus, $S=100$ is a relative benchmark, not a theoretical maximum, not ``100\% coupling,'' and not evidence that a stronger relationship could not exist in another set. Equal family weighting prevents the two FEDE descriptors from giving the distributional response twice the weight of the accuracy response. DPRS is therefore a relative within-study ranking rather than a universal effect size: scores from a different dataset or a different candidate set require recomputation of the normalizing maxima. Its practical role is to prioritize candidate directions when many potential auxiliary series are being compared. It does not replace class, raw magnitude, persistence fractions, or horizon profile. In particular, two directions can have similar DPRS values while belonging to different response classes and therefore have different scientific meanings.

\normalsize
\subsection{Sequential choice of the final empirical configuration}\label{sec:parameter_choice}

The final $W=400$, $m=14$, $H=14$ setting was reached sequentially rather than by varying many parameters simultaneously or choosing the configuration that maximized the final pairwise effects. The first broad FEDE screen used $W=200$, $m=5$, and $H=14$. The history length was then increased to $m=14$ while holding the window, horizon, model, and remaining settings fixed. Across aligned rolling windows, the ordinary log-RMSE response was comparatively stable between the two history settings (Pearson $r=0.856$), whereas the FEDE endpoint and early-slope responses were more sensitive ($r=0.310$ and $r=0.162$, respectively). This indicated that increasing history length alone did not resolve the visible variance of entropy-based descriptors.

The rolling window was therefore doubled to $W=400$ while retaining $m=14$ and $H=14$. This change increased the number of held-out forecast origins available for estimating each horizon-specific error distribution from 70 in the $W=200$, $m=14$ setting to 150 in the final setting. The choice reflects the same stability--resolution trade-off observed during synthetic FEDE calibration \citep{VelichkoFEDE2026}: more errors per horizon stabilize empirical distributions, but a longer window necessarily reduces temporal localization. The final configuration was selected for distributional stability, not because it maximized observed pairwise responses. The $W=200$ calculations are retained as a higher-resolution sensitivity analysis rather than treated as failed runs.

An earlier feasibility screen used shorter windows and a different predictor to identify representative Google/COVID pair types. It was not the final FEDE experiment and was not used to tune $W=400$ effects. The eight pairs in Table~\ref{tab:pairs} were fixed before the final regime-level run.

\normalsize
\subsection{Time-axis convention and interpretation of rolling estimates}\label{sec:time_axis}

The rolling calculation is causal with respect to the end of each window. An estimate stored at end date $T$ uses only the interval $[T-W+1,T]$. Therefore, a value obtained with $W=400$ is a property of an approximately 400-day historical regime and should not be interpreted as an effect occurring on a single calendar date.

For contextual visualization, the already-computed rolling values may be displayed at the center of their analysis interval,
\begin{equation}
 t_{\mathrm{center}}=\frac{t_{\mathrm{start}}+t_{\mathrm{end}}}{2},
\end{equation}
solely to facilitate comparison with the underlying source and target trajectories. This display shift does not alter a forecast, a train/test split, or the information available to the model, and therefore does not introduce future data into the calculation. Likewise, any moving averages used as visual guides in contextual plots are not used in LightGBM fitting, RMSE calculation, or FEDE estimation. The main statistical interpretation is consequently regime-level persistence rather than attribution of a metric peak to an individual epidemic wave, policy announcement, or calendar day.

\normalsize
\section{Results}\label{sec:results}

The rolling analysis produces a response that depends on both the location of the $W=400$ window and the forecast horizon. In particular, the underlying calculations contain horizon-specific quantities $g(t,h)$ and $d(t,h)$ for every admissible rolling window $t$ and every horizon $h=1,\ldots,14$. The results below deliberately use two complementary projections of this information. First, a temporal context plot retains the window axis while summarizing the horizon dependence through $G_{\log\RMSE}$, $D_{\FEDE}$, and $E_{\FEDE}$. Second, pair-level conclusions are based primarily on responses aggregated across the available rolling regimes, because the aim of this study is to identify clear dominant predictive patterns rather than to catalogue every local profile. Individual windows can therefore deviate from the aggregate pattern; this temporal heterogeneity is retained descriptively through positive-window fractions, interquartile ranges, and the complete context plots in the Appendix. Behavioral and epidemiological interpretation is deferred to Section~\ref{sec:discussion}.

\normalsize
\subsection{Temporal context of the rolling calculation}\label{sec:results_temporal}

Figure~\ref{fig:temporal_example} shows the complete context plot for B1 (``covid symptoms''$\leftrightarrow$hospitalized). The upper panel displays the two standardized observed series using 14-day visual means, while the three lower panels show the two explicit predictive directions, ``covid symptoms''$\rightarrow$hospitalized and hospitalized$\rightarrow$``covid symptoms''. This representation is useful because it makes the geometry of the $W=400$ analysis explicit. Each metric point summarizes one complete 400-day interval but is plotted at the center of that interval for visual comparison with the observed series. Consequently, the metric curves begin later and end earlier than the raw series, leaving visible margins of approximately half a window at the two ends. These margins are therefore a consequence of the wide rolling window and centered display, not missing calculations or an additional trimming step.

The B1 trajectories also illustrate why local maxima and minima are not treated as event-specific findings. Adjacent estimates share 399 of 400 days, and extended positive and negative intervals coexist over the record. For B1, ``covid symptoms''$\rightarrow$hospitalized is positive in 56.8\% of windows for mean log-RMSE gain, 56.9\% for the FEDE endpoint response, and 66.6\% for the FEDE early-slope response; hospitalized$\rightarrow$``covid symptoms'' is positive in 28.4\%, 40.1\%, and 48.9\% of windows, respectively. These fractions describe temporal persistence, not independent replications. Full context plots for the remaining seven pairs are provided in Appendix~\ref{app:data_pipeline}.

\par\medskip
\refstepcounter{figure}\label{fig:temporal_example}
\begin{center}
\includegraphics[width=0.96\textwidth]{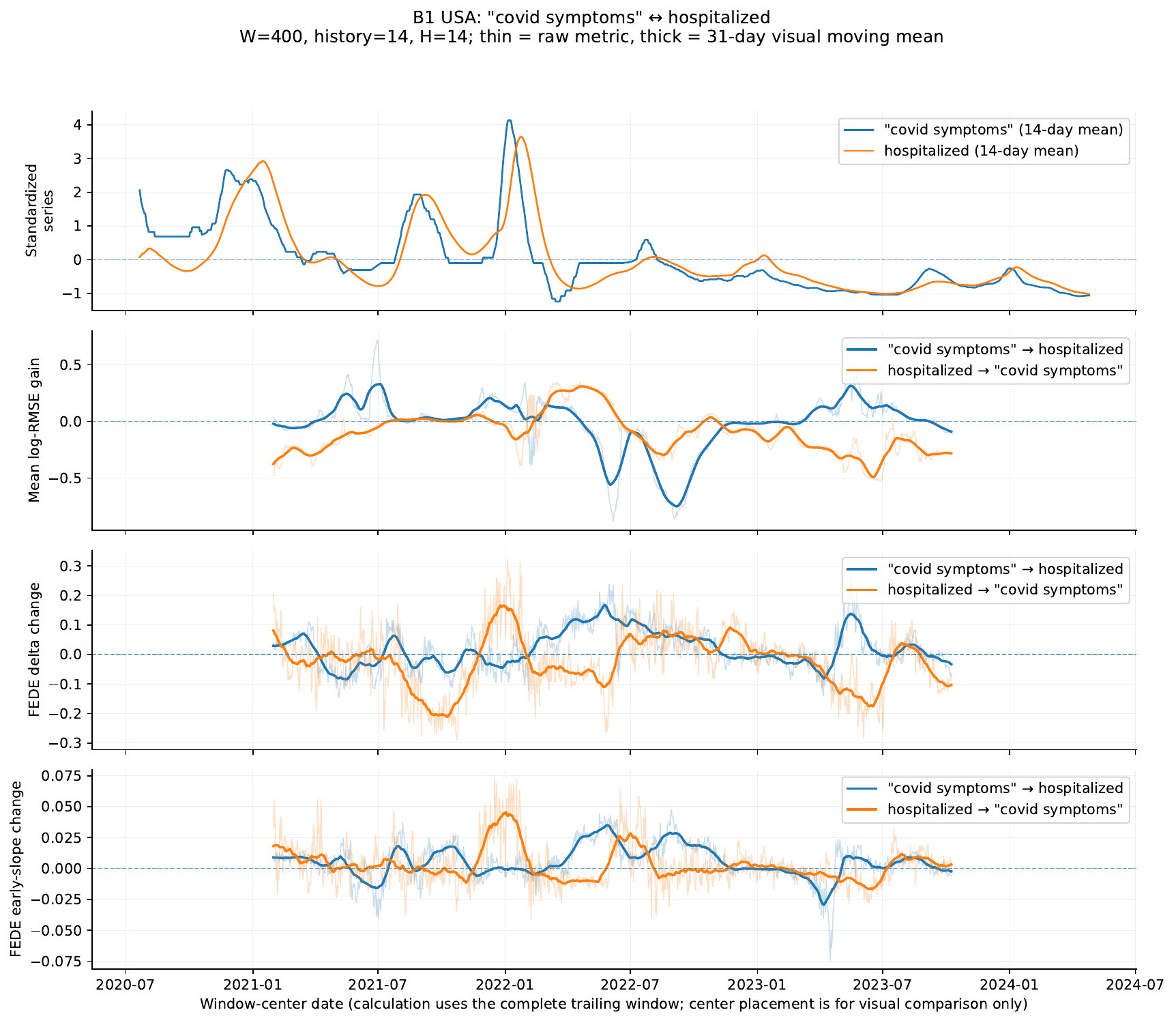}
\end{center}
\noindent\small\textbf{Figure \thefigure.} Temporal context of B1, ``covid symptoms''$\leftrightarrow$hospitalized. The top panel shows the two observed daily series as standardized 14-day visual means. The lower panels show rolling mean log-RMSE gain, FEDE endpoint response, and FEDE early-slope response for ``covid symptoms''$\rightarrow$hospitalized and hospitalized$\rightarrow$``covid symptoms'', as identified directly in the legends. Each response point is calculated from a complete $W=400$ window and displayed at the window center; the resulting left and right margins are therefore expected consequences of the centered wide-window representation. Thin curves are raw one-day rolling estimates and thick curves are 31-day moving means used only as visual guides.\par\medskip
\normalsize

\normalsize
\subsection{Aggregate bidirectional response landscape}\label{sec:results_overview}

For pair-level inference, we next aggregate the rolling responses rather than assigning a separate interpretation to every local regime. Table~\ref{tab:directional_results} summarizes the mean response and the fraction of rolling windows with positive response for every pair and direction, while Fig.~\ref{fig:overall_landscape} displays the corresponding mean values. Table~\ref{tab:pair_classification} then translates those numerical summaries, together with the dominant horizon profiles where needed, into the first-level response classes defined in Table~\ref{tab:response_matrix}. The purpose of this aggregation is not to assert that a pair is temporally invariant. Instead, it identifies relationships for which a recognizable directional effect remains after combining many overlapping annual-scale regimes.

The resulting map separates clear class assignments from directions that remain mixed. B7 provides the clearest Class I (concordant) search $\rightarrow$ indicator case. B2--B4 show the most pronounced metric-dependent asymmetry because their opposite directions switch between Class II (RMSE-only) and Class III (FEDE-only). B1 shows a moderate but heterogeneous Class III response and B5 a low-persistence Class III response in the search $\rightarrow$ indicator direction, whereas their reverse directions are mixed rather than forced into a class. B6 is mixed in both directions because the RMSE and the two FEDE summaries do not form a coherent first-level sign pattern. B8 falls into Class III in both directions, but its role as a seasonal/nonspecific stress case prevents a source-specific interpretation. No direction is used as a clean empirical Class IV example; mixed status should not be read as proof of no relationship.

The two FEDE descriptors are more mutually coherent than either is with RMSE gain. Across the eight pairs analysed in both directions, the mean within-direction Pearson correlation between FEDE endpoint and early-slope responses is $0.481$ (median $0.506$). The corresponding mean correlations of RMSE gain with FEDE endpoint and FEDE early slope are $-0.243$ and $-0.205$, respectively. These summary correlations reinforce the class map: FEDE responses often move together, but they need not track conventional accuracy gain.

\par\medskip
\refstepcounter{table}\label{tab:directional_results}
\noindent\textbf{Table \thetable}\\[-1mm]
\noindent\small Mean rolling responses for all eight pairs. F denotes search $\rightarrow$ paired indicator and R denotes paired indicator $\rightarrow$ search. Values in parentheses are the fractions of rolling windows in which the corresponding response is positive. $G=G_{\log\RMSE}$, $D=D_{\FEDE}$, and $E=E_{\FEDE}$.\par\smallskip
\scriptsize
\begin{tabularx}{\textwidth}{@{}p{0.45cm}YYYYYY@{}}
\toprule
ID & F $G$ & R $G$ & F $D$ & R $D$ & F $E$ & R $E$ \\
\midrule
B1 & $-0.017$ (0.57) & $-0.098$ (0.28) & $0.019$ (0.57) & $-0.023$ (0.40) & $0.006$ (0.67) & $0.003$ (0.49) \\
B2 & $-0.068$ (0.18) & $0.019$ (0.54) & $0.039$ (0.72) & $-0.026$ (0.31) & $0.006$ (0.69) & $-0.002$ (0.43) \\
B3 & $0.104$ (0.64) & $-0.018$ (0.35) & $-0.018$ (0.40) & $0.048$ (0.66) & $-0.003$ (0.41) & $0.001$ (0.46) \\
B4 & $0.026$ (0.58) & $-0.146$ (0.40) & $-0.005$ (0.45) & $0.042$ (0.53) & $-0.003$ (0.34) & $0.006$ (0.56) \\
B5 & $-0.094$ (0.24) & $-0.184$ (0.24) & $0.016$ (0.51) & $-0.037$ (0.36) & $0.003$ (0.53) & $0.001$ (0.50) \\
B6 & $0.016$ (0.31) & $-0.017$ (0.22) & $0.004$ (0.37) & $0.016$ (0.44) & $-0.001$ (0.34) & $-0.002$ (0.38) \\
B7 & $0.009$ (0.73) & $-0.204$ (0.40) & $0.040$ (0.72) & $0.001$ (0.53) & $0.003$ (0.62) & $0.006$ (0.66) \\
B8 & $-0.139$ (0.35) & $-0.219$ (0.23) & $0.017$ (0.58) & $0.006$ (0.55) & $0.001$ (0.56) & $0.002$ (0.59) \\
\bottomrule
\end{tabularx}\par\medskip
\normalsize

\par\medskip
\refstepcounter{table}\label{tab:pair_classification}
\noindent\textbf{Table \thetable}\\[-1mm]
\noindent\small First-level classification of all eight pairs in both directions. Classes I--IV are defined in Table~\ref{tab:response_matrix} and describe response type, not effect size. ``Mixed'' is used when the FEDE summaries disagree or when the dominant profile does not support a stable first-level assignment.\\[1mm]
\scriptsize
\begin{tabularx}{\textwidth}{@{}p{0.45cm}p{3.55cm}p{3.55cm}Y@{}}
\toprule
ID & Search $\rightarrow$ paired indicator & Paired indicator $\rightarrow$ search & Pair-level reading \\
\midrule
B1 & \textbf{Class III} (FEDE-only); positive FEDE means with non-positive mean RMSE gain & \textbf{Mixed}; endpoint FEDE is negative while early-slope FEDE is near zero/slightly positive & Forward direction shows a persistent distributional tendency, especially in the early-slope descriptor; reverse direction lacks a coherent positive first-level pattern \\
B2 & \textbf{Class III} (FEDE-only) & \textbf{Class II} (RMSE-only) & Clear metric-dependent class switch \\
B3 & \textbf{Class II} (RMSE-only) & \textbf{Class III} (FEDE-only) & Strongest metric-dependent bidirectional asymmetry \\
B4 & \textbf{Class II} (RMSE-only) & \textbf{Class III} (FEDE-only) & Forward accuracy response is horizon-limited; reverse response is distributional \\
B5 & \textbf{Class III} (FEDE-only), but low persistence & \textbf{Mixed}; negative endpoint FEDE and near-zero early-slope FEDE & World-level attenuation; no stable reverse class \\
B6 & \textbf{Mixed}; small positive RMSE and endpoint FEDE means but negative early-slope FEDE & \textbf{Mixed}; non-positive RMSE, positive endpoint FEDE, negative early-slope FEDE & No stable first-level class in either direction \\
B7 & \textbf{Class I} (concordant) & \textbf{Class III} (FEDE-only) & Clearest concordant forward response; reverse direction lacks RMSE gain \\
B8 & \textbf{Class III} (FEDE-only) & \textbf{Class III} (FEDE-only) & Nonspecific/seasonal stress case; distributional response is not treated as source-specific evidence \\
\bottomrule
\end{tabularx}
\normalsize\par\medskip

\par\medskip
\refstepcounter{figure}\label{fig:overall_landscape}
\begin{center}
\includegraphics[width=0.95\textwidth]{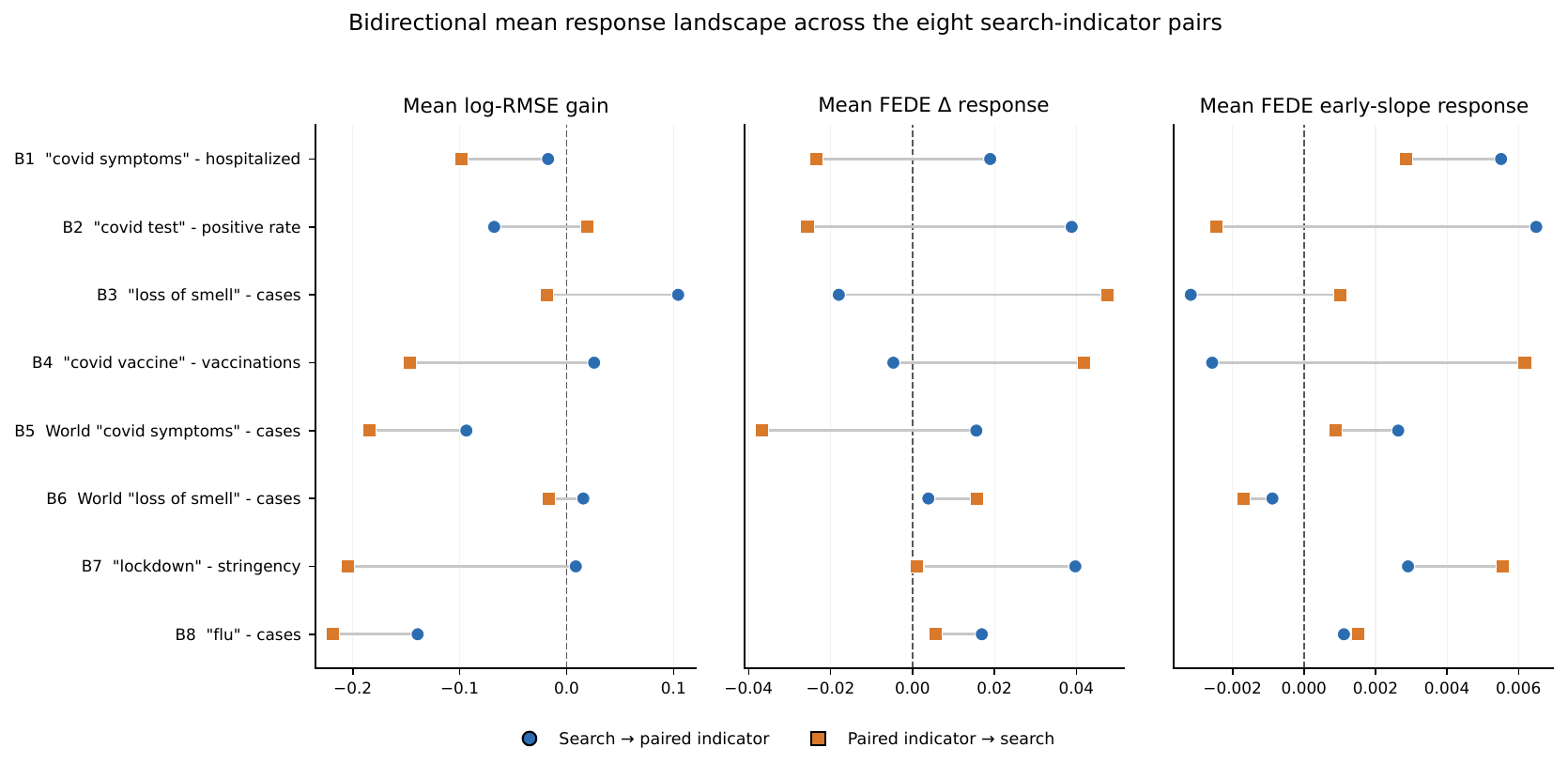}
\end{center}
\noindent\small\textbf{Figure \thefigure.} Aggregate bidirectional response landscape across the eight pre-specified search--indicator pairs. Circles denote search $\rightarrow$ paired indicator and squares paired indicator $\rightarrow$ search. The three panels report the mean rolling log-RMSE gain, FEDE endpoint response, and FEDE early-slope response. The zero line separates positive and negative mean responses. These means summarize the dominant response over the available rolling regimes and do not imply temporal invariance.\par\medskip
\normalsize

\normalsize
\subsection{Dominant horizon-resolved profiles}\label{sec:results_horizon}

The scalar means in Table~\ref{tab:directional_results} further compress the forecast-horizon dimension. To recover that structure while retaining the pair-level focus, we average the horizon-specific quantities over all rolling windows,
\begin{equation}
 \bar g(h)=\frac{1}{N}\sum_{t=1}^{N}g(t,h),\qquad
 \bar d(h)=\frac{1}{N}\sum_{t=1}^{N}d(t,h),
\end{equation}
where $N$ is the number of complete $W=400$ windows available for the corresponding pair. Figure~\ref{fig:horizon_profiles} shows these window-averaged profiles for B2, B3, B4, and B7, which were retained because their aggregate shapes are distinct and directly interpretable. The shaded regions show the 25th--75th percentile range across rolling windows. They are descriptive interquartile ranges, not confidence intervals, because adjacent windows are highly dependent. Thus, the solid line represents the dominant profile after aggregation, whereas the shading makes explicit that individual rolling regimes can depart from that profile.

\par\medskip
\refstepcounter{figure}\label{fig:horizon_profiles}
\begin{center}
\includegraphics[width=0.97\textwidth]{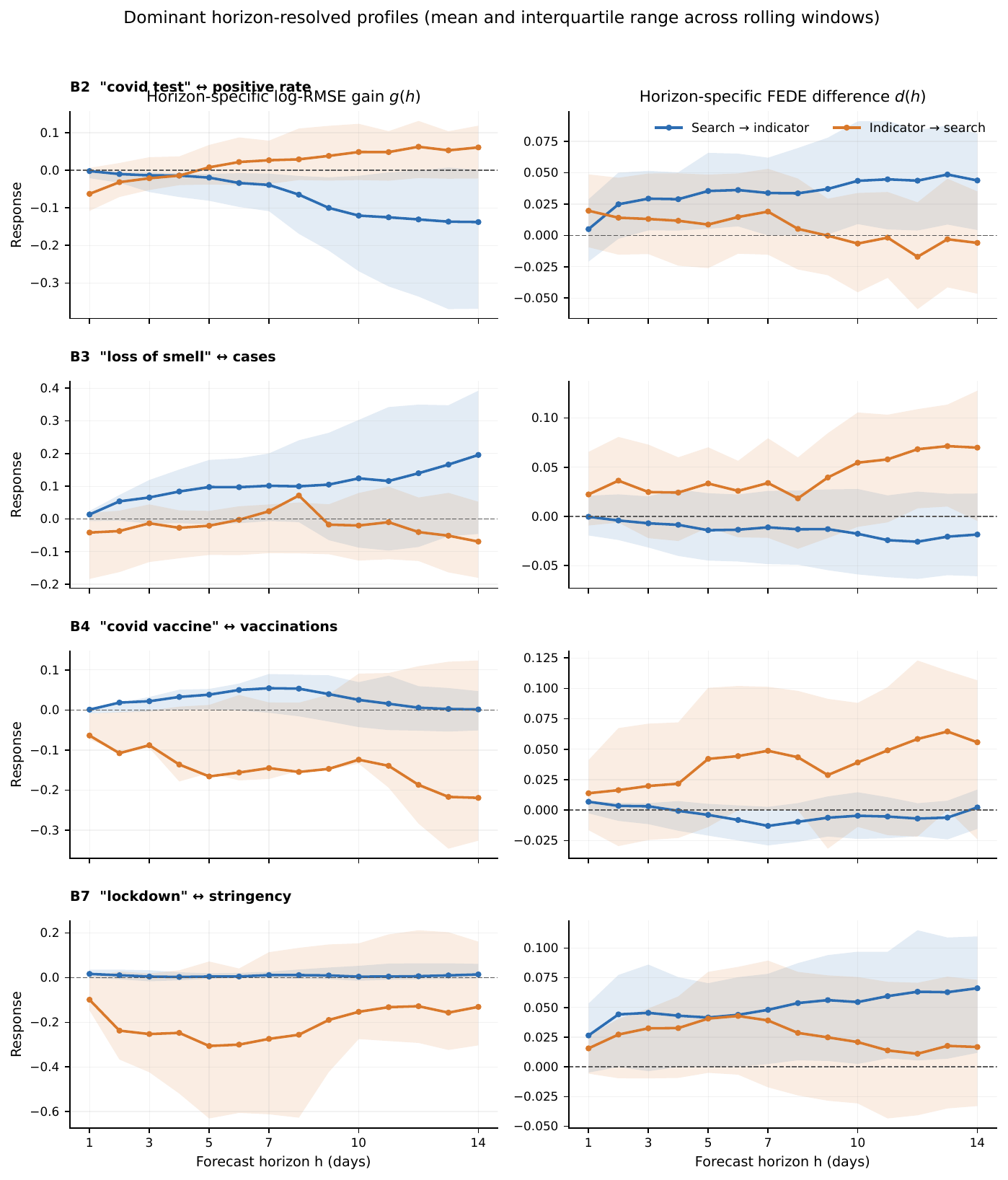}
\end{center}
\noindent\small\textbf{Figure \thefigure.} Dominant horizon-resolved profiles for four representative bidirectional relationships. Solid curves show the mean across all available rolling windows of horizon-specific log-RMSE gain $g(t,h)$ (left) and FEDE difference $d(t,h)=\FEDE_{\mathrm{aux}}(t,h)-\FEDE_{\mathrm{base}}(t,h)$ (right). Blue denotes search $\rightarrow$ paired indicator and orange denotes paired indicator $\rightarrow$ search: B2, ``covid test''$\leftrightarrow$positive rate; B3, ``loss of smell''$\leftrightarrow$cases; B4, ``covid vaccine''$\leftrightarrow$vaccinations; and B7, ``lockdown''$\leftrightarrow$stringency. Shaded regions show the interquartile range across rolling windows and quantify temporal heterogeneity; they are not confidence intervals. Both directions use the same forecasting protocol.\par\medskip
\normalsize

B2 exhibits the clearest divergence between point-accuracy and distributional responses. In the ``covid test''$\rightarrow$positive-rate calculation, $\bar g(h)$ changes from approximately $-0.002$ at $h=1$ to $-0.138$ at $h=14$, while $\bar d(h)$ rises from $0.005$ to $0.044$. Positive rate$\rightarrow$``covid test'' develops a positive RMSE gain at intermediate horizons and reaches approximately $0.061$ at $h=14$, whereas its FEDE difference is close to zero or negative at the longest horizon.

B3 displays a different directional split. For ``loss of smell''$\rightarrow$cases, $\bar g(h)$ increases from $0.014$ at $h=1$ to $0.196$ at $h=14$, while $\bar d(h)$ remains negative. For cases$\rightarrow$``loss of smell'', $\bar d(h)$ increases from approximately $0.022$ to $0.070$ between $h=1$ and $h=14$, whereas the RMSE gain does not remain positive across the full horizon range. B4 shows a finite-horizon accuracy response for ``covid vaccine''$\rightarrow$vaccinations: $\bar g(h)$ reaches approximately $0.054$ near $h=7$ but returns close to zero by $h=14$. Vaccinations$\rightarrow$``covid vaccine'' has negative RMSE gain throughout the shown range while $\bar d(h)$ grows from about $0.014$ to $0.056$. Finally, ``lockdown''$\rightarrow$stringency retains a modest positive RMSE gain across the horizon range while its FEDE difference increases from about $0.026$ at $h=1$ to $0.066$ at $h=14$; the stringency$\rightarrow$``lockdown'' RMSE profile remains negative.

Viewed through the pair-level classification in Table~\ref{tab:pair_classification} (with class definitions in Table~\ref{tab:response_matrix}), these dominant profiles separate naturally. ``lockdown''$\rightarrow$stringency is Class I (concordant response), the closest real-data analogue of the synthetic-like pattern. ``covid test''$\rightarrow$positive rate is Class III (FEDE-only response): FEDE is positive while RMSE gain is non-positive. ``loss of smell''$\rightarrow$cases is Class II (RMSE-only response), whereas cases$\rightarrow$``loss of smell'' is Class III. B4 shows the same directional class switch: ``covid vaccine''$\rightarrow$vaccinations is Class II, whereas vaccinations$\rightarrow$``covid vaccine'' is Class III. These assignments are only the first level of interpretation; the detailed shapes in Fig.~\ref{fig:horizon_profiles} provide the second level by showing how each response develops across $h=1,\ldots,14$. The B3 and B4 switches from Class II to Class III are therefore direct examples of metric-dependent bidirectional asymmetry.

These aggregate profiles are not intended as an exhaustive taxonomy of all possible local shapes. The complete rolling calculations contain a broader variety of increasing, decreasing, approximately flat, and non-monotonic profiles. The present analysis deliberately emphasizes pairs for which averaging across many broad regimes leaves a clear dominant horizon dependence; a systematic classification of local profile morphologies would constitute a separate extension rather than a requirement for the present conclusions.

\normalsize
\subsection{Representative forecast-error distributions}\label{sec:results_distributions}

The aggregate profiles identify which type of response dominates a pair, but RMSE and FEDE still summarize different properties of the same test-set forecasts. Figure~\ref{fig:error_distributions} therefore returns to the underlying signed errors for three representative windows to illustrate what the two summaries can mean at the distribution level. In this figure, \emph{baseline} means the target-history-only forecast, whereas \emph{augmented} means the otherwise matched forecast after adding the paired-series history; the arrows in the panel annotations therefore read baseline $\rightarrow$ augmented. These examples are explanatory rather than a second basis for pair-level inference. Windows were selected algorithmically rather than by visual peak picking: among windows satisfying the relevant sign pattern, the displayed window is the case closest to the median three-component response vector after standardization. Baseline and augmented errors use the same 16-bin range within each selected window, matching the FEDE calculation.

The B2 example represents a Class III (FEDE-only) response: over 19 July 2020--22 August 2021, $G_{\log\RMSE}=-0.109$, $D_{\FEDE}=0.072$, and $E_{\FEDE}=0.0129$. The B3 example represents a Class II (RMSE-only) response: over 15 July 2021--18 August 2022, $G_{\log\RMSE}=0.313$, while $D_{\FEDE}=-0.0597$ and $E_{\FEDE}=-0.0124$. The B7 example represents a Class I (concordant) response: over 18 June 2020--22 July 2021, all three summaries are positive ($G_{\log\RMSE}=0.0346$, $D_{\FEDE}=0.0722$, $E_{\FEDE}=0.00892$). The distributions at $h=1$, 7, and 14 show that a reduced RMSE and a changed FEDE trajectory correspond to visibly different modifications of the held-out error distribution.

\par\medskip
\refstepcounter{figure}\label{fig:error_distributions}
\begin{center}
\includegraphics[width=0.98\textwidth]{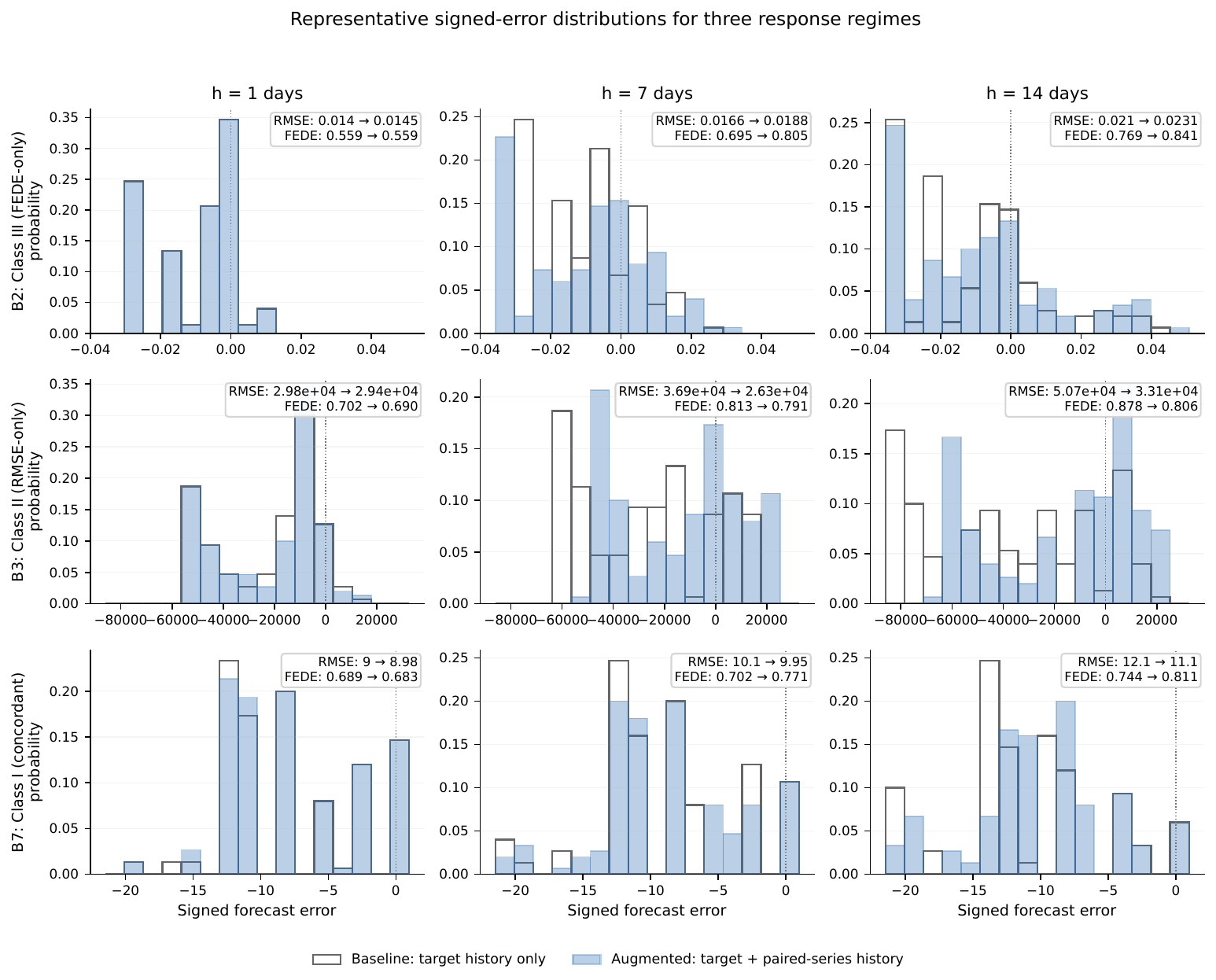}
\end{center}
\noindent\small\textbf{Figure \thefigure.} Forecast-error distribution anatomy of three representative response regimes. Rows show a Class III (FEDE-only) B2 window, a Class II (RMSE-only) B3 window, and a Class I (concordant) B7 window; columns show forecast horizons $h=1$, 7, and 14 days. Baseline denotes the target-history-only model and augmented denotes the matched model with paired-series history added. Numeric annotations show baseline $\rightarrow$ augmented values. Within each selected window, both branches use the same forecast origins and the same 16-bin error range. Windows were chosen algorithmically as the cases closest to the median response vector among all windows satisfying the corresponding sign pattern, not as visually selected extrema.\par\medskip
\normalsize

For a direct numerical reading of Fig.~\ref{fig:error_distributions}, Table~\ref{tab:fig5_numeric} reports the three displayed horizons. RMSE change is expressed as
\[
I_{\RMSE}(h)=100\,\frac{\RMSE_b(h)-\RMSE_a(h)}{\RMSE_b(h)},
\]
so positive values denote improvement and negative values denote worsening after adding the paired series. FEDE is already a normalized entropy on $[0,1]$, so its change is reported in absolute units, $\Delta\FEDE(h)=\FEDE_a(h)-\FEDE_b(h)$, rather than as a percentage ``improvement.'' The response class itself is assigned from the window-level summaries $G_{\log\RMSE}$, $D_{\FEDE}$, and $E_{\FEDE}$ across all 14 horizons; the three rows per case merely show how the underlying quantities evolve within that window.

\par\medskip
\refstepcounter{table}\label{tab:fig5_numeric}
\noindent\textbf{Table \thetable}\\[-1mm]
\noindent\small Numerical interpretation of the three representative windows in Fig.~\ref{fig:error_distributions}. RMSE change is positive for improvement and negative for worsening. $\Delta\FEDE$ is the augmented-minus-baseline normalized entropy difference.\\[1mm]
\scriptsize
\begin{tabularx}{\textwidth}{@{}p{2.65cm}c p{2.45cm}p{1.45cm}p{2.35cm}p{1.25cm}@{}}
\toprule
Case / class & $h$ & RMSE baseline $\rightarrow$ augmented & RMSE change & FEDE baseline $\rightarrow$ augmented & $\Delta\FEDE$ \\
\midrule
B2 ``covid test''$\rightarrow$positive rate, Class III & 1 & 0.01400 $\rightarrow$ 0.01453 & $-3.7\%$ & 0.559 $\rightarrow$ 0.559 & $0.000$ \\
 & 7 & 0.01659 $\rightarrow$ 0.01884 & $-13.6\%$ & 0.695 $\rightarrow$ 0.805 & $+0.111$ \\
 & 14 & 0.02104 $\rightarrow$ 0.02309 & $-9.7\%$ & 0.769 $\rightarrow$ 0.841 & $+0.072$ \\
\addlinespace
B3 ``loss of smell''$\rightarrow$cases, Class II & 1 & 29,816 $\rightarrow$ 29,420 & $+1.3\%$ & 0.702 $\rightarrow$ 0.690 & $-0.012$ \\
 & 7 & 36,925 $\rightarrow$ 26,334 & $+28.7\%$ & 0.813 $\rightarrow$ 0.791 & $-0.022$ \\
 & 14 & 50,697 $\rightarrow$ 33,149 & $+34.6\%$ & 0.878 $\rightarrow$ 0.806 & $-0.072$ \\
\addlinespace
B7 ``lockdown''$\rightarrow$stringency, Class I & 1 & 8.997 $\rightarrow$ 8.980 & $+0.2\%$ & 0.689 $\rightarrow$ 0.683 & $-0.005$ \\
 & 7 & 10.103 $\rightarrow$ 9.952 & $+1.5\%$ & 0.702 $\rightarrow$ 0.771 & $+0.069$ \\
 & 14 & 12.112 $\rightarrow$ 11.136 & $+8.1\%$ & 0.744 $\rightarrow$ 0.811 & $+0.067$ \\
\bottomrule
\end{tabularx}
\normalsize\par\medskip

Taken together, the results support a deliberately compact pair-level interpretation. Temporal context confirms that individual annual-scale regimes vary. Aggregation then identifies which directional and horizon-dependent effects remain clear over the available record, while the error-distribution examples demonstrate why an aggregate RMSE response and an aggregate FEDE response need not carry the same information. The subsequent Discussion interprets only these dominant patterns rather than assigning a separate mechanism to every rolling window.

\normalsize
\section{Discussion}\label{sec:discussion}

\normalsize
\subsection{From controlled nonlinear coupling to observational predictive structure}\label{sec:discussion_synthetic}

The main result of this study is not that Google Trends variables are universally useful predictors of COVID-19 indicators. That proposition would be too broad and is already known to depend strongly on query choice, geography, epidemic phase, and evaluation design \citep{Ma2022Review,Saegner2022,Rovetta2021}. The more specific result is that, once the same bidirectional forecasting experiment is applied to observational data, conventional forecast-error gain and FEDE no longer behave as interchangeable measures of one underlying effect. This contrasts with the controlled setting in which FEDE was introduced \citep{VelichkoFEDE2026}. In those synthetic benchmarks, the source--target interaction was imposed by a known coupling channel, the direction of that channel was controlled, and its strength could be varied independently of unrelated social or epidemiological mechanisms. Under those conditions, stronger coupling tended to produce a common response in forecast-error magnitude and in the horizon evolution of the error distribution.

The Google Trends--COVID setting is structurally different. A search series can respond to private symptoms, public case reports, media attention, policy announcements, testing availability, vaccination campaigns, or a mixture of these processes. The epidemiological and policy series are themselves affected by reporting practices and interventions. Thus, there is no single scalar coupling parameter whose increase should force all forecast diagnostics to move together. The observed divergence is therefore a substantive result rather than an anomaly. Across the eight pairs analysed in both directions, the two FEDE descriptors are moderately associated with each other, whereas their associations with RMSE gain are weak and negative on average. This indicates that mean forecast-error magnitude and the horizon-dependent structure of signed errors capture related but non-equivalent properties of out-of-sample prediction.

This distinction also clarifies the role of the synthetic study. The earlier work established that FEDE is sensitive to nonlinear predictive coupling and robust to broad variations of noise and analysis parameters under controlled conditions. The present study does not repeat that calibration. Instead, it asks whether the same horizon-resolved description remains interpretable when the observed relationship may be bidirectional, regime dependent, and generated by several mechanisms at once. In this sense, the synthetic study establishes \emph{sensitivity}, whereas the present analysis addresses \emph{interpretability under observational complexity}.

Local signs are expected to fluctuate in finite observational windows, just as local forecast gains fluctuate in the synthetic benchmarks and become noisier when fewer forecast errors are available. The pair-level interpretation therefore does not rely on an isolated positive FEDE value or on a single rolling RMSE gain. It asks whether a dominant response direction survives aggregation across many broad $W=400$ regimes and whether that direction is supported by the corresponding horizon profile and positive-window persistence. A positive FEDE response is not assumed to be the only possible distributional change; it is the response direction that was coupling-consistent in the preceding synthetic calibration. Negative and mixed local FEDE responses are retained as temporal heterogeneity rather than relabeled as ``negative coupling.'' This distinction between noisy local responses and persistent aggregate signatures is also the motivation for the persistence weighting used in DPRS (Section~\ref{sec:dprs}).

\normalsize
\subsection{RMSE gain and FEDE describe different aspects of predictive contribution}\label{sec:discussion_metrics}

The empirical results support a practical distinction between two questions that are often conflated. The first is whether adding the source series reduces the average magnitude of the target forecast error. This is summarized here by $G_{\log\RMSE}$ and by the horizon-specific gain $g(h)$. The second is whether adding the source changes how the full signed-error distribution evolves as the prediction horizon increases. This is summarized by $D_{\FEDE}$, $E_{\FEDE}$, and the profile $d(h)$. A source can therefore be informative in the second sense without improving point accuracy in the first sense.

Figure~\ref{fig:concept_entropy} isolates this distinction in a deliberately simple artificial example. Two 30-point forecast scenarios are constructed with zero mean signed error and the same RMSE ($1.00$), yet their signed-error distributions differ sharply. Using the same normalized Shannon-entropy form as Eq.~\eqref{eq:fede} and common histogram bins, the concentrated two-level error pattern has normalized entropy $0.333$, whereas the broader error pattern has normalized entropy $0.993$. The example is conceptual rather than empirical and does not reproduce the full FEDE procedure by itself. Its purpose is narrower: identical average error magnitude can coexist with very different organization of signed errors. FEDE extends this principle by tracking such entropy across forecast horizons and by comparing the augmented and baseline branches.

\par\medskip
\refstepcounter{figure}\label{fig:concept_entropy}
\begin{center}
\includegraphics[width=0.93\textwidth]{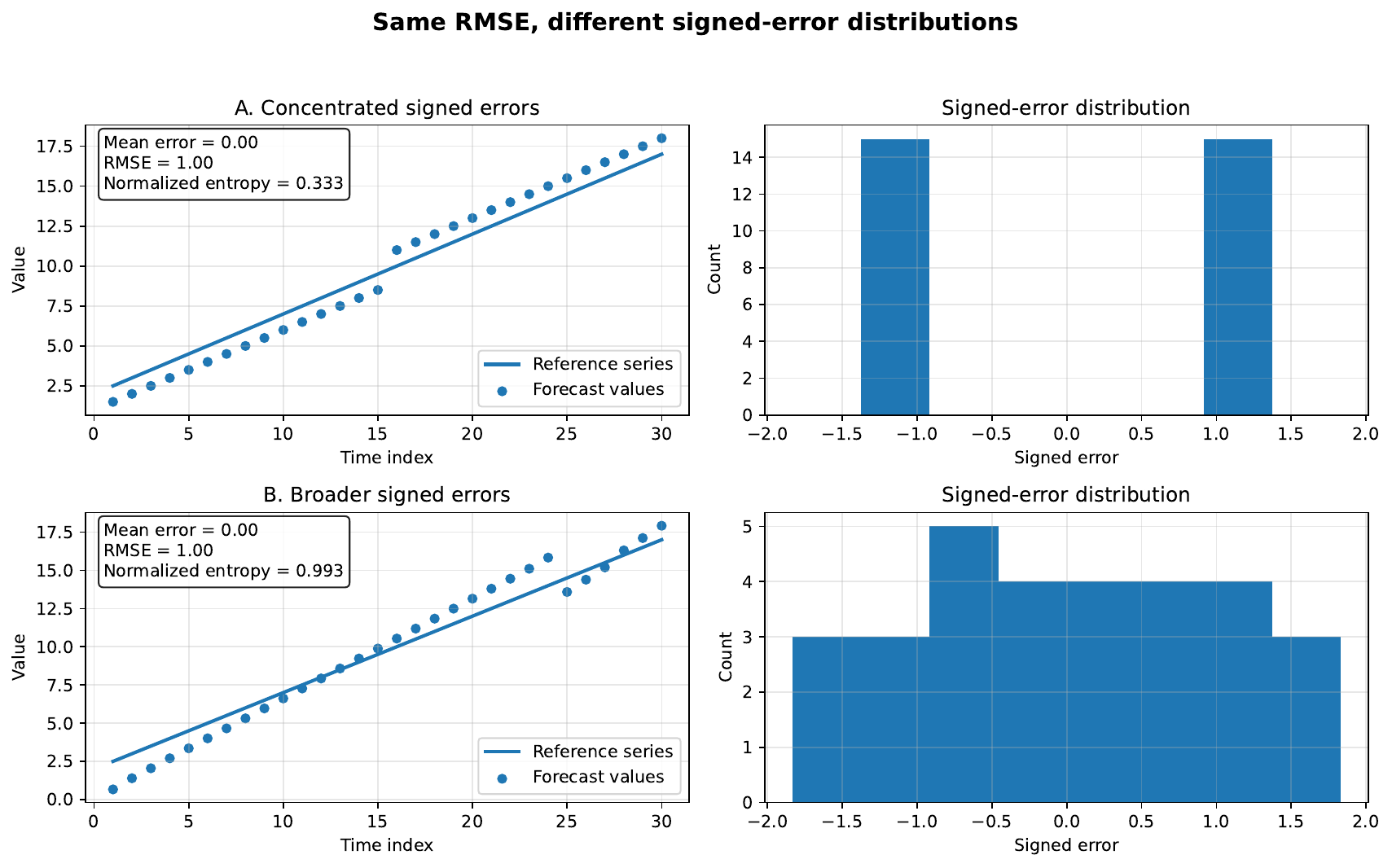}
\end{center}
\noindent\small\textbf{Figure \thefigure.} Conceptual illustration of identical RMSE with different signed-error distributions. Both artificial scenarios have zero mean signed error and RMSE $=1.00$, but their normalized Shannon entropies, computed in the same form as Eq.~\eqref{eq:fede} using common bin edges, are $0.333$ and $0.993$. The figure is an explanatory construction, not an empirical result and not a substitute for the horizon-resolved FEDE calculation.\par\medskip
\normalsize

B2 provides the clearest empirical example. For ``covid test''$\rightarrow$positive rate, the average RMSE response becomes increasingly negative toward longer horizons, but the mean FEDE difference becomes increasingly positive. The representative distributions in Fig.~\ref{fig:error_distributions} and their numerical summary in Table~\ref{tab:fig5_numeric} make the divergence explicit. At $h=7$, adding ``covid test'' worsens RMSE by $13.6\%$, yet the normalized FEDE increases by $0.111$; at $h=14$, RMSE is still $9.7\%$ worse while $\Delta\FEDE=+0.072$. Thus the auxiliary history changes the organization of signed errors across horizons without improving point accuracy. If the analysis were restricted to a single average accuracy score, this direction would largely be classified as unhelpful. FEDE instead identifies a reproducible horizon-dependent reorganization. Positive rate$\rightarrow$``covid test'' displays the complementary pattern, with a small conventional forecast gain but a much weaker FEDE response.

The opposite case is represented by B3. In the ``loss of smell''$\rightarrow$cases direction, the mean RMSE gain is the largest positive value in the study and increases strongly with horizon, whereas the FEDE response is weak or negative. In the representative B3 window, RMSE improves by $28.7\%$ at $h=7$ and $34.6\%$ at $h=14$, while $\Delta\FEDE$ remains negative ($-0.022$ and $-0.072$). Here the auxiliary series improves point prediction without producing the distributional signature observed in B2. B7 provides the concordant case: for ``lockdown''$\rightarrow$stringency, the representative window shows an $8.1\%$ RMSE improvement at $h=14$ together with $\Delta\FEDE=+0.067$. These three examples---Class III, Class II, and Class I, respectively---demonstrate that FEDE is best interpreted as complementary to conventional forecast error rather than as a replacement for it.

This interpretation is consistent with the methodological warnings in the Google Trends literature. Search data can contain real predictive information, but apparent associations can also be created or amplified by nonstationarity, media exposure, and changing public awareness \citep{Sato2021,SousaPinto2020,Rovetta2021}. A distribution-sensitive descriptor is therefore useful precisely because it does not assume that every relevant informational relationship must manifest as a uniform reduction in scalar forecast loss. At the same time, a positive FEDE response cannot by itself establish a specific behavioral or epidemiological mechanism; B8 provides an internal demonstration of this limitation.

\normalsize
\subsection{Bidirectional asymmetry across the eight search--indicator pairs}\label{sec:discussion_pairs}

The bidirectional design is central to the interpretation because every forward and reverse calculation uses the same model class, history length, horizon set, rolling window, and chronological validation protocol. Directional differences are therefore not artifacts of changing the forecasting engine. Throughout this discussion, Classes I--IV refer to the first-level sign-based response classes defined in Table~\ref{tab:response_matrix}, while Table~\ref{tab:pair_classification} shows the resulting assignment for all eight pairs in both directions. ``Mixed'' means that the sign-based summaries do not support a stable class; it does not mean that the two series are unrelated. Class labels describe response type rather than strength, which is judged separately from mean magnitude, positive-window persistence, interquartile spread, and the horizon profiles.

\paragraph{B1: symptom searches and hospitalization.}
B1 is best viewed as a consistency case rather than the main novelty of the study. Previous work has shown that symptom-related internet searches can precede hospital admissions and can improve short-horizon hospitalization forecasts, although the strength of the contribution varies across waves and regions \citep{Wang2022Hospital,Mellor2023Leading,Mellor2023Realtime,Wang2023Alert}. In Table~\ref{tab:pair_classification}, ``covid symptoms''$\rightarrow$hospitalized is Class III because its mean RMSE gain is non-positive while both FEDE summaries are positive, but its endpoint magnitude is below the clearest FEDE-only examples while its early-slope response is comparatively persistent. The reverse direction is mixed because the FEDE endpoint and early-slope summaries do not agree in sign. Thus B1 supports a modest forward distributional tendency rather than a uniformly positive accuracy effect. This is compatible with the literature's broader conclusion that symptom searches are useful but context dependent rather than universal leading indicators.

\paragraph{B2: testing interest and test positivity.}
B2 is the clearest Class III (FEDE-only) relation in the present set (Table~\ref{tab:response_matrix}). In the ``covid test''$\rightarrow$positive-rate direction, search history does not improve average RMSE; instead, the positive response is carried by FEDE through a systematic change in the signed-error distribution across forecast horizons. The reverse direction, positive rate$\rightarrow$``covid test'', is Class II (RMSE-only): it shows a modest conventional accuracy gain and weaker or negative FEDE response. A behavioral interpretation is possible but should remain cautious. Search interest can reflect intention to test, symptoms, concern, news exposure, or changes in testing access, while positive rate is itself a ratio influenced by who is being tested. The observed asymmetry is therefore more securely interpreted as evidence that the two series carry different forms of incremental predictive information than as evidence for a single causal chain from search to testing outcome.

\paragraph{B3: loss of smell and reported cases.}
B3 provides the strongest example of metric-dependent bidirectional asymmetry. At the first level (Table~\ref{tab:response_matrix}), ``loss of smell''$\rightarrow$cases is Class II (RMSE-only), whereas cases$\rightarrow$``loss of smell'' is Class III (FEDE-only). This pattern is particularly informative because the anosmia literature already supports both temporal directions. Searches for loss of smell have been reported to precede case increases by approximately one to three weeks in some settings \citep{Chen2022,Lippi2020Smell,Walker2020,Cherry2020}, whereas case surges and media exposure can stimulate later smell-related searching \citep{Higgins2020,Wu2020,Panuganti2020,SousaPinto2020,Madden2021}. Figure~\ref{fig:b3_pathways} summarizes two observational pathways that make these findings mutually compatible. In a symptom-led pathway, infection is followed by anosmia, symptom-related Google searching, subsequent testing or diagnosis, and eventual appearance in reported case counts. In an attention-led pathway, rising reported cases increase news coverage and public attention, raise awareness of anosmia, and stimulate later smell-related searching in a broader population. The first pathway is compatible with a search-to-cases leading signal, whereas the second is compatible with a cases-to-search response. Our forecasting results do not identify either pathway as the causal mechanism of a particular RMSE or FEDE change. They show instead that the two directions can be separated within one matched design and can be expressed through different forecast-error properties. This is stronger than treating apparently opposite lead--lag findings from separate studies as mutually exclusive.

\par\medskip
\refstepcounter{figure}\label{fig:b3_pathways}
\begin{center}
\includegraphics[width=0.92\textwidth]{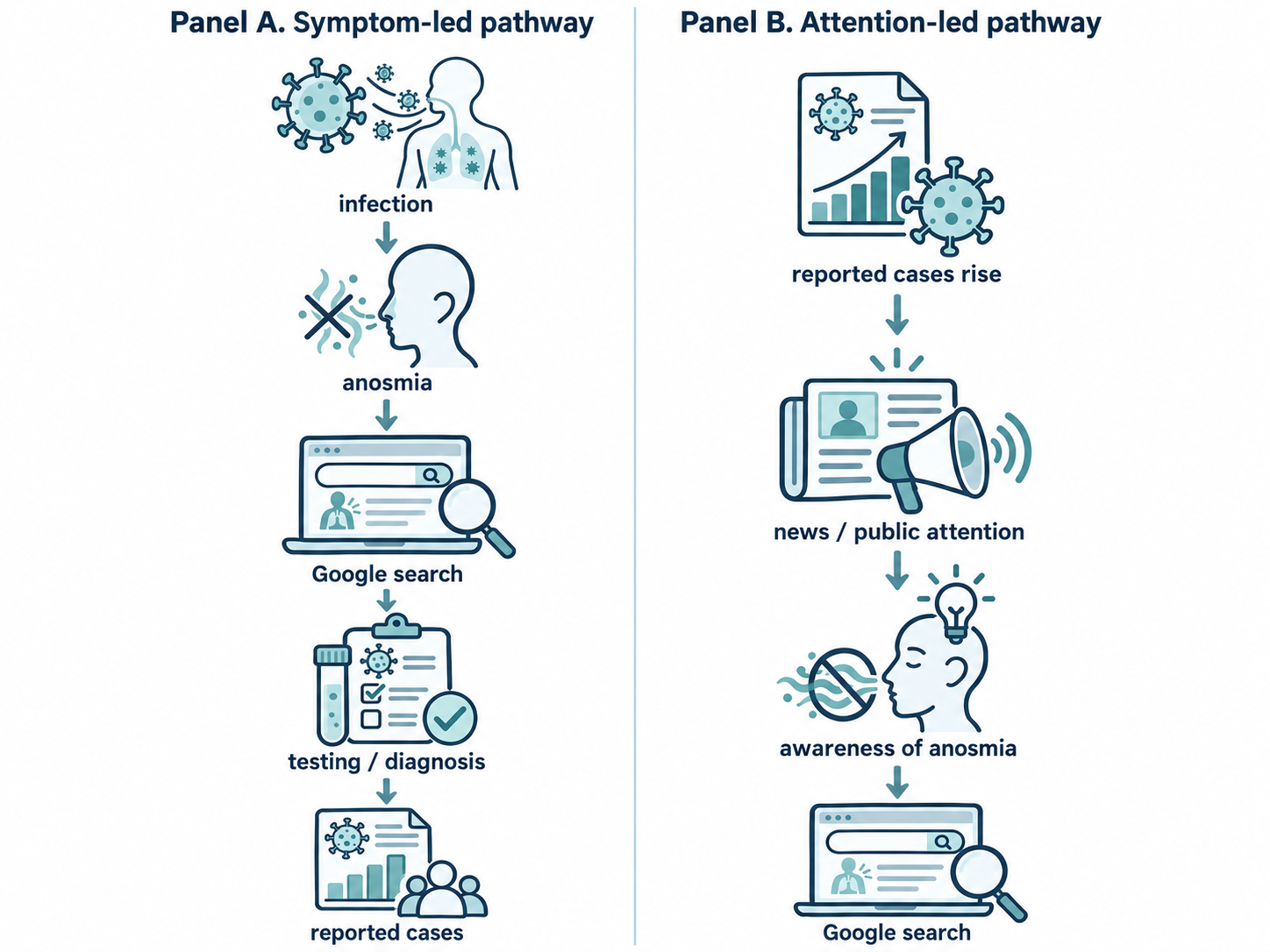}
\end{center}
\noindent\small\textbf{Figure \thefigure.} Two observational pathways consistent with opposite temporal directions in the loss-of-smell literature. Panel A shows a symptom-led pathway in which infection and anosmia can precede symptom-related Google searching, testing or diagnosis, and later inclusion in reported case counts. Panel B shows an attention-led pathway in which rising reported cases can increase news and public attention, raise awareness of anosmia, and stimulate subsequent Google searching. The diagram is an interpretive synthesis of mechanisms discussed in the literature, not a causal identification result from the present forecasting analysis.\par\medskip
\normalsize

\paragraph{B4: vaccine information and vaccination activity.}
B4 exhibits another Class II $\leftrightarrow$ Class III switch (Table~\ref{tab:response_matrix}). In the ``covid vaccine''$\rightarrow$vaccinations direction (Class II, RMSE-only), the accuracy response is concentrated over a finite horizon range. The mean RMSE gain rises from near zero, reaches its maximum around horizons of six to eight days, and then decays toward zero by $h=14$. Vaccinations$\rightarrow$``covid vaccine'' is Class III (FEDE-only): it has negative RMSE gain but a positive FEDE response that grows toward longer horizons. The forward profile is compatible with an information-to-action delay in which search activity may precede a later behavioral outcome, while the reverse distributional response is compatible with population search behavior reacting to vaccination activity or the information environment surrounding it. Neither mechanism is directly observed here, so the appropriate conclusion is the narrower one: the relation is horizon dependent and the two directions are expressed through different forecast-error properties.

\paragraph{B5--B6: world-level generalization.}
The two world-level symptom/case pairs are weaker and less persistent than the strongest USA relations, but they should not simply be labeled ``no relationship.'' Table~\ref{tab:pair_classification} assigns B5 search$\rightarrow$cases to Class III because the mean FEDE responses are positive while RMSE gain is negative, yet the positive-window fractions are close to one half and the response magnitude is small; the reverse direction is mixed. B6 is mixed in both directions because the RMSE and the two FEDE summaries do not form a stable sign pattern. These results caution against assuming that a search-based signal transfers unchanged across spatial aggregation. Published Google Trends studies have repeatedly reported geographical and phase-dependent variability in predictive performance \citep{Husnayain2021,Fulk2021,Marty2023,Asplin2025}. In the present setting, a worldwide series additionally combines countries with asynchronous epidemic waves, different testing and reporting systems, different levels of internet use, and different linguistic relationships to the English query terms. The data therefore support a limited generalization claim: the clearest USA-level profiles do not automatically survive worldwide aggregation. The study does not determine which of the above sources of heterogeneity is responsible for the attenuation.

\paragraph{B7: lockdown search and policy stringency.}
B7 is the clearest Class I (concordant) forward case in Table~\ref{tab:pair_classification}. The ``lockdown''$\rightarrow$stringency calculation has positive mean responses for all three descriptors and positive responses in a large fraction of windows. The reverse direction, stringency$\rightarrow$``lockdown'', is Class III: its mean RMSE gain is strongly negative while the FEDE summaries are positive but much smaller in the endpoint component. Importantly, B7 shows that divergence between RMSE and FEDE is not inevitable. When a relation has a sufficiently coherent predictive structure, the conventional and distributional summaries can reinforce one another. Search behavior around restrictions may plausibly contain anticipatory information about policy discussion, public expectations, or impending measures, while implemented restrictions can also provoke search activity; the present analysis does not separate these mechanisms. Its robust conclusion is that ``lockdown''$\rightarrow$stringency is substantially more coherent across metrics than the reverse direction.

\paragraph{B8: seasonal/nonspecific stress case.}
B8 plays a different role. Table~\ref{tab:pair_classification} places both directions in Class III (FEDE-only) because mean RMSE gain is negative for both ``flu''$\rightarrow$cases and cases$\rightarrow$``flu'', whereas the FEDE summaries are positive. The class label identifies the response type, not its specificity or causal meaning. The result is therefore not treated as evidence that ``flu'' searches provide a specific predictive channel for COVID-19 cases. Instead, B8 demonstrates that a distributional response can arise in a pair sharing seasonal, respiratory-symptom, and information-environment structure even when point prediction does not improve. This observation is consistent with broader warnings that Google Trends signals can reflect common public attention or shared temporal structure rather than a unique disease process \citep{Sato2021,SousaPinto2020,Rovetta2021}. B8 consequently acts as an internal interpretive safeguard: FEDE detects a change in error-distribution dynamics, but additional validation is required before such a change can be attributed to a source-specific mechanism.

\normalsize
\subsection{Forecast horizon is part of the relationship, not only a tuning parameter}\label{sec:discussion_horizon}

The horizon-resolved profiles add information that is lost both in a single lead--lag coefficient and in a single averaged forecast score. Published studies already show that the predictive value of search data can be concentrated at short or intermediate horizons and can weaken as the horizon increases \citep{Wang2022Hospital,Chen2022,Amusa2022,Yang2023,Radziwon2023}. Our results extend this observation by showing that different forecast diagnostics can change in different directions over the same horizon range. B2 becomes increasingly unfavorable by RMSE while its FEDE difference becomes increasingly positive. For B3, ``loss of smell''$\rightarrow$cases gains point-prediction value with increasing horizon while the FEDE response for cases$\rightarrow$``loss of smell'' strengthens. B4 has an intermediate-horizon RMSE maximum, and B7 remains comparatively concordant over the full range.

These patterns motivate a more precise language for observational predictive relationships. Instead of stating only that ``$A$ predicts $B$,'' it is more informative to ask what additional information $A$ contributes to $B$, at which horizons, and through which property of the forecast errors. A source may improve near-term point estimates, alter longer-horizon uncertainty structure, or do both. The distinction is particularly relevant for information-behavior data because different delays can coexist: symptom recognition, information seeking, diagnostic confirmation, policy response, and public reaction need not operate on the same time scale.

The forecast horizon $h$ should also not be confused with a conventional lead or lag obtained by shifting two observed series against one another. Here, $h$ is the distance between the information available at a forecast origin and the future target being predicted. A maximum at $h=7$ therefore means that the auxiliary history provides its strongest model-relative incremental information for a seven-day-ahead forecast under this design; it does not establish that an observed search event causes a downstream event exactly seven days later.

\normalsize
\subsection{Why regime-level aggregation is appropriate here}\label{sec:discussion_aggregation}

The raw calculations are two-dimensional in time and horizon: every rolling window has its own $g(t,h)$ and $d(t,h)$ profiles. The contextual trajectories and appendix figures show that local regimes are heterogeneous, and the interquartile bands in Fig.~\ref{fig:horizon_profiles} make this heterogeneity explicit. The goal of the present article, however, is not to classify every local shape. Such a taxonomy could include increasing, decreasing, flat, U-shaped, and inverted-U profiles for both RMSE and FEDE and would constitute a separate methodological problem.

Instead, the analysis asks whether a recognizable profile survives aggregation across many broad regimes. This choice follows directly from the use of $W=400$. Each point is estimated from an approximately annual window and contains 150 held-out errors at each horizon, improving the stability of empirical error distributions at the cost of local time resolution. Averaging over these windows then identifies the dominant horizon dependence of a pair over its available record. The corresponding interquartile range is retained to show how much local variation lies around that dominant response. Because neighboring windows overlap by 399 days, neither the mean profile nor the positive-window fraction should be interpreted as a collection of independent replications or as a formal significance test.

This regime-level viewpoint also prevents overinterpretation of visual coincidence. The centered context plots are useful for seeing how the observed source and target series relate to the rolling metrics, but a metric peak is supported by a complete 400-day interval rather than by the calendar date at which the point is drawn. The present results therefore support statements about persistent or dominant predictive structure, not claims that a particular epidemic peak or policy announcement ``caused'' a local FEDE excursion.

\normalsize
\subsection{Implications for information behavior and digital epidemiology}\label{sec:discussion_information}

From an information-science perspective, the results suggest that web-search data should not be treated only as noisy proxies for epidemiological variables. Search activity is itself a behavioral response generated within the same system being monitored. It can precede formal observations, respond to those observations, or reflect common external information. The bidirectional design makes these possibilities visible: an epidemiological variable can improve prediction of search activity even when the reverse direction carries a different kind of predictive contribution. This is particularly clear in B2--B4, where the two directions are not simply weaker and stronger versions of the same response.

Accordingly, the practical value of search data is not exhausted by asking whether they reduce RMSE. A search series may carry useful information about a changing predictive regime even when it does not improve a point forecast. Conversely, a strong reduction in RMSE does not guarantee a positive FEDE response. The complete map in Table~\ref{tab:pair_classification} also shows why response type and response strength should be reported separately: the same Class III label covers both clear examples such as B2 and lower-persistence or nonspecific examples such as B5 and B8. For digital surveillance and information-behavior studies, reporting class, magnitude, persistence, and horizon profile together can therefore distinguish whether the auxiliary information primarily improves forecast magnitude, reorganizes the horizon-dependent distribution of errors, or produces a concordant response. When many candidate auxiliary series are available, relative DPRS adds a practical screening layer by ordering these signatures within the fixed candidate set while leaving the directly interpretable RMSE effect and FEDE components visible. The same framework is applicable beyond COVID-19 whenever two aligned observational time series may participate in a feedback-rich information system.

\normalsize
\subsection{A practical directional predictive signature}\label{sec:discussion_signature}
The practical output of the framework is best viewed as a \emph{directional predictive signature} rather than a binary statement that two series are or are not ``connected.'' For each direction, the signature contains (i) the first-level response class, (ii) the equivalent RMSE effect, (iii) the signed FEDE endpoint and early-slope responses, (iv) persistence of each positive response across broad rolling regimes, (v) the dominant horizon behavior, and (vi) the secondary relative DPRS ranking from Section~\ref{sec:dprs}. The forward and reverse signatures are reported separately; their difference is itself part of the result. This form is deliberately practical: when an analyst has a set of candidate auxiliary series, the RMSE-effect column answers the conventional point-forecast question, $D/E$ describes the distributional response, the positive-window fractions indicate temporal persistence, the horizon profile shows where the contribution is expressed, and relative DPRS supplies a single ordering within that candidate set.

Table~\ref{tab:predictive_signature} gives this practical summary for all 16 directed calculations and orders them by relative DPRS. The ranking is informative precisely because it does not equate response type with strength. B3 ``loss of smell''$\rightarrow$cases has the highest relative DPRS in the current set ($S=100$) and is accuracy-dominant, whereas the reverse B3 direction is FEDE-dominant and has $S=42$. B2 ``covid test''$\rightarrow$positive rate is nearly as high in the within-study ordering ($S=98$) but belongs to Class III rather than Class II; its score comes from a persistent FEDE response rather than improved RMSE, while the reverse B2 direction has only $S=5$. B7 ``lockdown''$\rightarrow$stringency combines both response families and has $S=78$, making it the clearest concordant signature; the reverse direction is FEDE-only with $S=38$. B4 illustrates another form of asymmetry: the forward RMSE-only signature has $S=15$, whereas the reverse FEDE-only signature has $S=22$.

The lower part of the ranking is equally informative. B5 and both B6 directions have little persistent positive strength after world-level aggregation, consistent with attenuation of the clearer USA patterns. B8 produces low but nonzero FEDE-based scores in both directions ($S=11$ and $S=7$ after rounding), yet its nonspecific seasonal role prevents that distributional response from being interpreted as source-specific predictive evidence. Thus, Table~\ref{tab:predictive_signature} can be used as a screening table: sort candidate directions by relative DPRS, then read the class, RMSE effect, FEDE components, persistence, and horizon behavior to determine what produced that position in the ranking. The score should therefore be used as a navigation aid through the multidimensional signature, not as a replacement for it, and $S=100$ should never be read as an absolute upper bound on possible predictive strength.

\par\medskip
\refstepcounter{table}\label{tab:predictive_signature}
\noindent\textbf{Table \thetable}\\[-1mm]
\noindent\small Directional predictive signatures for all eight pairs in both directions, sorted by relative persistence-weighted directional predictive response strength (DPRS). ``RMSE effect'' is $100[1-\exp(-\bar G)]$; positive values denote equivalent RMSE improvement and negative values worsening. FEDE is reported as $\bar D/\bar E$. Persistence gives the percentage of positive windows for $G/D/E$. Relative DPRS is a within-study 0--100 ranking from Eq.~\eqref{eq:dprs}; $100$ denotes the strongest observed direction in this comparison set, not a theoretical maximum, significance statistic, or causal coupling coefficient.\par\smallskip
{\scriptsize
\renewcommand{\arraystretch}{1.08}
\begin{longtable}{@{}p{0.45cm}p{2.90cm}p{0.90cm}p{1.15cm}p{1.75cm}p{1.35cm}p{0.68cm}p{3.35cm}@{}}
\toprule
Rank & Direction & Class & RMSE effect & FEDE $D/E$ & Positive windows $G/D/E$ & Rel. DPRS & Dominant reading \\
\midrule
\endfirsthead
\toprule
Rank & Direction & Class & RMSE effect & FEDE $D/E$ & Positive windows $G/D/E$ & Rel. DPRS & Dominant reading \\
\midrule
\endhead
1 & B3 ``loss of smell''$\rightarrow$cases & II & $+9.9\%$ & \mbox{$-0.018/-0.003$} & 64/40/41\% & 100 & Strong persistent accuracy response; gain increases with $h$ \\
2 & B2 ``covid test''$\rightarrow$positive rate & III & $-7.0\%$ & \mbox{$+0.039/+0.006$} & 18/72/69\% & 98 & Strong persistent FEDE response while RMSE worsens \\
3 & B7 ``lockdown''$\rightarrow$stringency & I & $+0.9\%$ & \mbox{$+0.040/+0.003$} & 73/72/62\% & 78 & Clearest concordant response; FEDE grows with $h$ \\
4 & B1 ``covid symptoms''$\rightarrow$hospitalized & III & $-1.8\%$ & \mbox{$+0.019/+0.006$} & 57/57/67\% & 45 & Persistent forward distributional tendency, strongest in early slope \\
5 & B3 cases$\rightarrow$``loss of smell'' & III & $-1.8\%$ & \mbox{$+0.048/+0.001$} & 35/66/46\% & 42 & Endpoint FEDE response without accuracy gain \\
6 & B7 stringency$\rightarrow$``lockdown'' & III & $-22.7\%$ & \mbox{$+0.001/+0.006$} & 40/53/66\% & 38 & Early FEDE response; no positive RMSE contribution \\
7 & B4 vaccinations$\rightarrow$``covid vaccine'' & III & $-15.8\%$ & \mbox{$+0.042/+0.006$} & 40/53/56\% & 22 & Positive FEDE means but limited sign persistence \\
8 & B4 ``covid vaccine''$\rightarrow$vaccinations & II & $+2.5\%$ & \mbox{$-0.005/-0.003$} & 58/45/34\% & 15 & Accuracy gain peaks at intermediate horizons \\
9 & B8 ``flu''$\rightarrow$cases & III & $-15.0\%$ & \mbox{$+0.017/+0.001$} & 35/58/56\% & 11 & Nonspecific seasonal distributional response \\
10 & B8 cases$\rightarrow$``flu'' & III & $-24.5\%$ & \mbox{$+0.006/+0.002$} & 23/55/59\% & 7 & Reverse nonspecific distributional response \\
11 & B2 positive rate$\rightarrow$``covid test'' & II & $+1.9\%$ & \mbox{$-0.026/-0.002$} & 54/31/43\% & 5 & Small RMSE-only response \\
12 & B5 World ``covid symptoms''$\rightarrow$cases & III & $-9.8\%$ & \mbox{$+0.016/+0.003$} & 24/51/53\% & 3 & Weakly persistent world-level FEDE tendency \\
13 & B5 World cases$\rightarrow$``covid symptoms'' & Mixed & $-20.2\%$ & \mbox{$-0.037/+0.001$} & 24/36/50\% & 0 & No stable reverse world-level response \\
14 & B1 hospitalized$\rightarrow$``covid symptoms'' & Mixed & $-10.3\%$ & \mbox{$-0.023/+0.003$} & 28/40/49\% & 0 & No persistent positive first-level signature \\
15 & B6 World cases$\rightarrow$``loss of smell'' & Mixed & $-1.7\%$ & \mbox{$+0.016/-0.002$} & 22/44/38\% & 0 & Mixed FEDE signs and no persistent positive response \\
16 & B6 World ``loss of smell''$\rightarrow$cases & Mixed & $+1.5\%$ & \mbox{$+0.004/-0.001$} & 31/37/34\% & 0 & Positive means do not persist in a majority of regimes \\
\bottomrule
\end{longtable}
}
\normalsize

\normalsize
\subsection{Defensible claims and boundaries of interpretation}\label{sec:discussion_claims}

The empirical findings support four claims that are narrower than a causal claim but stronger than a simple correlation statement:
\begin{enumerate}[label=(\arabic*)]
\item \textbf{Forecast-accuracy gain and FEDE are complementary rather than redundant.} In the observational pairs studied here, a source can reduce average forecast error, reorganize the horizon-dependent signed-error distribution, or do both. The conceptual entropy example in Fig.~\ref{fig:concept_entropy}, the systematic B2--B3 contrast, and the representative distributions quantified in Table~\ref{tab:fig5_numeric} make this distinction directly observable.
\item \textbf{Bidirectional predictive coupling can be metric dependent.} Opposite directions of the same pair need not be expressed through the same forecast property. B3 is the clearest case, with point-accuracy gain dominating one direction and FEDE response dominating the other; B2 and B4 show related asymmetries.
\item \textbf{Forecast horizon is part of the predictive relation.} The sign and magnitude of incremental information can change across $h=1,\ldots,14$, so a single aggregate error score or a single ``best lag'' can conceal meaningful structure. B2, B3, and B4 provide distinct horizon-dependent examples, whereas B7 demonstrates a more concordant profile.
\item \textbf{Aggregated regime-level patterns should not be extrapolated to individual time points or interpreted as evidence of causality.} The $W=400$ design and time aggregation identify dominant patterns that survive broad temporal variation, while the context plots and interquartile ranges document substantial local heterogeneity. Common forcing, media attention, seasonality, policy, and reporting dynamics remain possible explanations for part of the observed coupling.
\end{enumerate}

These four points define the scientific scope of the paper's central contribution. Operationally, they are brought together in the directional predictive signature introduced in Section~\ref{sec:discussion_signature} and summarized in Table~\ref{tab:predictive_signature}. For each ordered pair, the signature reports response class, RMSE effect, FEDE endpoint and early-slope responses, temporal persistence, and horizon dependence, while relative DPRS provides a secondary within-study ranking. Thus, the framework produces not only a qualitative statement about whether two series exhibit predictive coupling, but a structured profile of how that coupling is expressed, how persistent it is, at which horizons it appears, and how its positive response compares with the other candidate directions in the same screening set.

The analysis does not claim that Google search activity causes the corresponding epidemiological or policy variable, nor that a positive FEDE response is by itself evidence of source-specific causation. Rather, it shows that a matched, bidirectional, multi-horizon forecasting experiment can separate several forms of incremental predictive information that would be collapsed by a single scalar association or forecast-error measure. This distinction provides the conceptual bridge from the controlled FEDE benchmarks to real information-behavior data and is the principal result to be carried into the final conclusions.

\normalsize
\section{Limitations}\label{sec:limitations}
Several limitations are intrinsic to the design. First, adjacent $W=400$ windows overlap by 399 days, so rolling trajectories are highly dependent and fractions of positive windows quantify persistence rather than independent replication. Second, the long window deliberately exchanges temporal localization for more stable empirical error distributions; the method therefore cannot assign a predictive effect to an individual epidemic peak or policy event. Third, Google Trends values are relative, normalized, and sensitive to query formulation, public awareness, platform sampling, and media exposure. The chunk-stitching procedure preserves daily resolution and internal temporal comparability but cannot convert relative search interest into absolute population behavior. Fourth, pairwise predictive contribution can reflect common forcing, policy, seasonality, reporting practices, or unmeasured media dynamics; reverse-direction asymmetry reduces but does not eliminate these ambiguities. Fifth, all conclusions are conditional on the chosen history length, LightGBM model class, and observational variables. Sixth, DPRS is normalized to the 16 directions in the present dataset and is therefore a descriptive relative ranking rather than a universal effect size, significance statistic, or causal coupling coefficient. Its value of 100 marks the strongest observed direction in the current comparison set only; adding or removing candidate directions can change all nonzero scores because the normalizing maxima must be recomputed. The present work therefore uses predictive rather than causal language.

A focused empirical null analysis using temporally constrained surrogate sources would be a useful next step for distinguishing source-specific contribution from slow common dynamics and seasonality, particularly for control-like pairs. Such a test is not included in the present iteration and no surrogate-based significance claim is made here.

\normalsize
\section{Conclusions}\label{sec:conclusion}
This study transferred the FEDE predictive-coupling framework from controlled nonlinear benchmarks \citep{VelichkoFEDE2026} to a substantially more complex observational setting in which Google search behavior, epidemiological dynamics, public policy, media exposure, and behavioral feedback may coexist. The synthetic setting provided a comparatively simple calibration: one imposed source--target channel tended to move conventional forecast gain and FEDE responses in the same direction. The Google Trends--COVID-19 analysis shows that this one-dimensional picture does not generally survive in observational data. Instead, improvement in forecast accuracy and reorganization of the signed-error distribution across horizons can separate and can be expressed differently in opposite predictive directions.

The eight pre-specified pairs do not all provide equally clear responses. The pair-level map separates clean Class I--III cases from mixed directions without treating the classes as an effect-size scale. The practical directional signatures make the distinction quantitative within the present comparison set: B3 ``loss of smell''$\rightarrow$cases has the highest relative DPRS ($100$) and is accuracy-dominant, B2 ``covid test''$\rightarrow$positive rate is nearly as high in the relative ranking but FEDE-dominant (DPRS $=98$), and B7 ``lockdown''$\rightarrow$stringency is the clearest concordant direction (DPRS $=78$). Their reverse scores are different (42, 5, and 38, respectively), illustrating that both response type and relative position in the candidate set are directional. B4 also switches class, but its forward and reverse scores are more modest (15 and 22). B1 retains a moderate forward distributional signature, B5--B6 show substantial world-level attenuation, and B8 remains a low-ranked nonspecific seasonal stress case rather than evidence of a source-specific channel. None of these scores should be read as a percentage of an absolute coupling maximum.

Taken together, the analysis supports four defensible conclusions. First, \textbf{forecast-accuracy gain and FEDE are complementary rather than redundant}: a candidate source may reduce average error, alter the horizon-dependent error distribution, or do both. Second, \textbf{bidirectional predictive coupling can be metric dependent}: opposite directions of one observed pair need not be weaker and stronger versions of the same response. Third, \textbf{forecast horizon is part of the predictive relation}: one aggregate score or one selected lag can conceal where incremental information strengthens, weakens, or changes character. Fourth, \textbf{aggregated regime-level patterns should not be extrapolated to individual time points or interpreted as evidence of causality}: the $W=400$ design identifies dominant patterns that survive broad temporal variation, while overlapping windows, shared external drivers, media attention, seasonality, policy, and reporting dynamics preclude local causal attribution.

The principal contribution is therefore not another claim that Google Trends ``predicts COVID-19,'' but a more differentiated account of what predictive contribution can mean in a feedback-rich observational system. The practical realization is a directional predictive profile: for each ordered pair, the framework returns response class, directly interpretable RMSE effect, FEDE endpoint and early-slope effects, temporal persistence, horizon dependence, and a secondary relative persistence-weighted DPRS ranking. This profile can identify whether a candidate auxiliary series is useful mainly for point prediction, mainly for detecting a change in error-distribution dynamics, for both, or for neither in a persistent way; it also indicates the forecast horizons over which that contribution is expressed. In practical screening problems, the analyst can therefore compare a set of candidate signals in one table, rank the directions by relative DPRS, and then use the remaining signature components to understand why a candidate ranks where it does. DPRS provides a simple within-set ordering when one number is needed, but the class and profile remain essential because similar scalar scores can arise from qualitatively different forecast-error mechanisms. This distinction provides a compact bridge between controlled FEDE calibration and real information-behavior data and offers a general strategy for other paired observational time series in which information seeking and the monitored process may influence one another over multiple timescales.

\section*{Data and code availability}
The reusable FEDE implementation and controlled synthetic benchmark suite are available through the citable software record associated with the methodological study \citep{VelichkoFEDESoftware2026}. The complete article-specific reproducibility package for the present Google Trends--COVID analysis is archived on Zenodo \citep{VelichkoGoogleCOVID2026}, DOI \href{https://doi.org/10.5281/zenodo.23499353}{10.5281/zenodo.23499353}. It contains the Google Trends acquisition and harmonization scripts, OWID preparation scripts, frozen compact input tables, pair definitions, complete analysis configurations, postprocessing utilities, the directional-signature/DPRS calculator and reference CSV, the conceptual RMSE--entropy figure generator, and article-scale reference results. The supplied Windows launcher reproduces the final calculations and performs numerical verification against the archived reference outputs. Appendix~\ref{app:data_pipeline} summarizes the upstream data-construction workflow and archive contents.

\section*{Funding}
This research was supported by the Russian Science Foundation (grant no. 22-11-00055-P).

\appendix
\normalsize
\section{Data acquisition, harmonization, and reproducibility archive}\label{app:data_pipeline}

The main FEDE workflow in Fig.~\ref{fig:main_workflow} assumes that two aligned daily series have already been obtained. Figure~\ref{fig:data_pipeline} documents the upstream construction of those inputs. Google Trends and OWID are prepared independently and are joined only after each side has been converted into a daily series. The 250-day Google Trends chunks with 45-day overlap belong exclusively to this acquisition stage: they are used to recover a long daily relative-interest series despite interval-dependent Google Trends normalization and must not be confused with the $W=400$ rolling windows used later for predictive analysis. No FEDE, RMSE gain, or forward/reverse comparison is computed during this appendix pipeline.

\setcounter{figure}{0}
\renewcommand{\thefigure}{A.\arabic{figure}}
\renewcommand{\theHfigure}{A.\arabic{figure}}
\par\medskip
\refstepcounter{figure}\label{fig:data_pipeline}
\begin{center}
\includegraphics[width=0.95\textwidth]{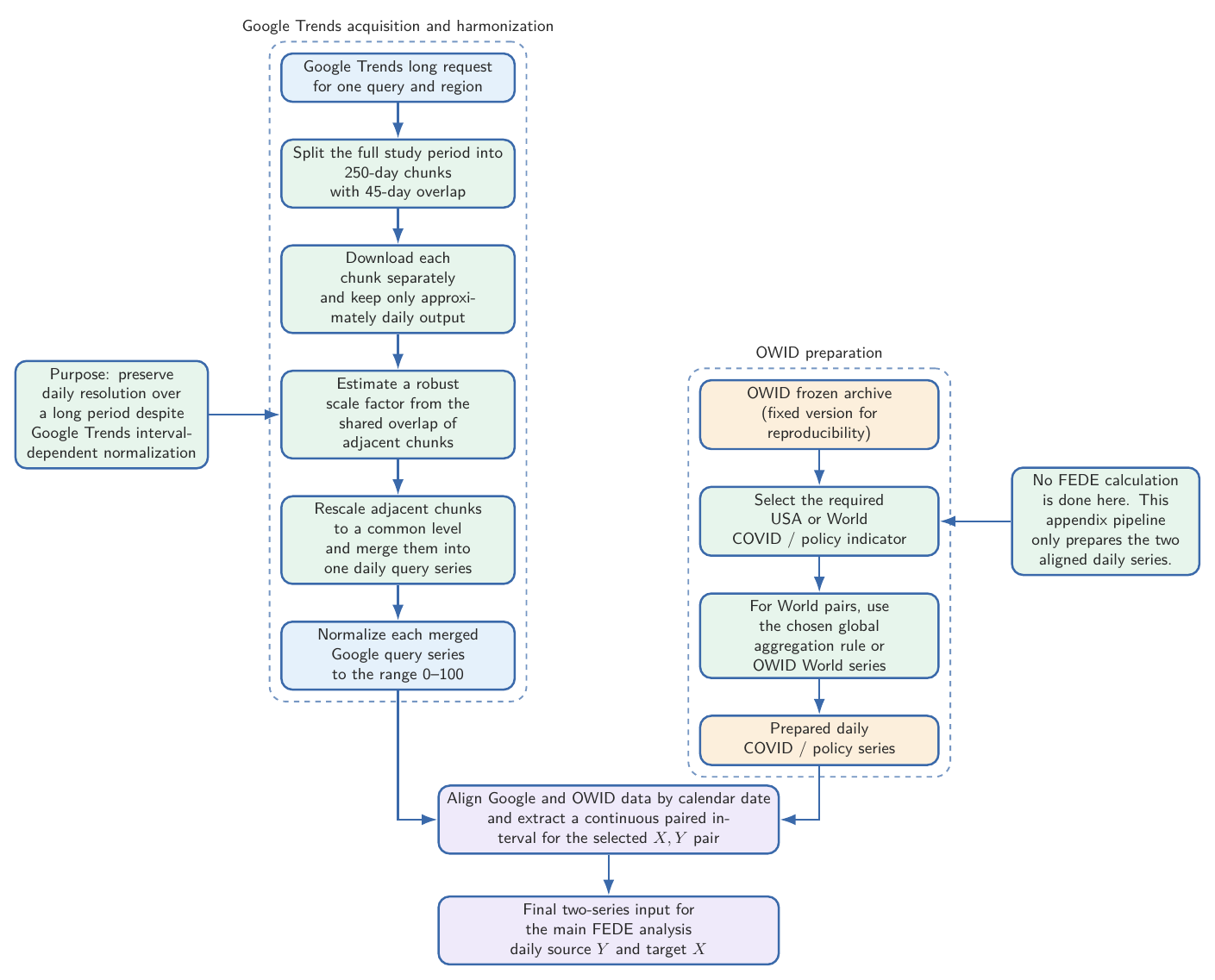}
\end{center}
\noindent\small\textbf{Figure \thefigure.} Upstream data-acquisition and harmonization pipeline. Long Google Trends series are reconstructed from overlapping daily-resolution requests, rescaled through their shared overlaps, merged, and normalized within each query. COVID-19 and policy indicators are prepared independently from a fixed OWID archive. The two branches are then aligned by calendar date and reduced to a continuous paired interval for the selected source $Y$ and target $X$. This stage prepares the inputs for the main FEDE experiment but does not itself perform predictive-coupling analysis.\par\medskip
\normalsize

The complete temporal context plots for the remaining seven search--indicator pairs are collected below for transparency; B1 is shown at full size in the main text (Fig.~\ref{fig:temporal_example}). Each appendix panel includes the standardized 14-day visual means of the search series and its paired epidemiological/policy indicator above the three response trajectories for the two explicit directions. Raw rolling metrics are shown by thin curves and 31-day visual moving means by thick curves. These figures document temporal heterogeneity but are not assigned separate mechanistic interpretations in the present study.

\par\medskip
\refstepcounter{figure}\label{fig:context_b2_b3}
\noindent\begin{minipage}[t]{0.495\textwidth}\centering
\includegraphics[width=\linewidth]{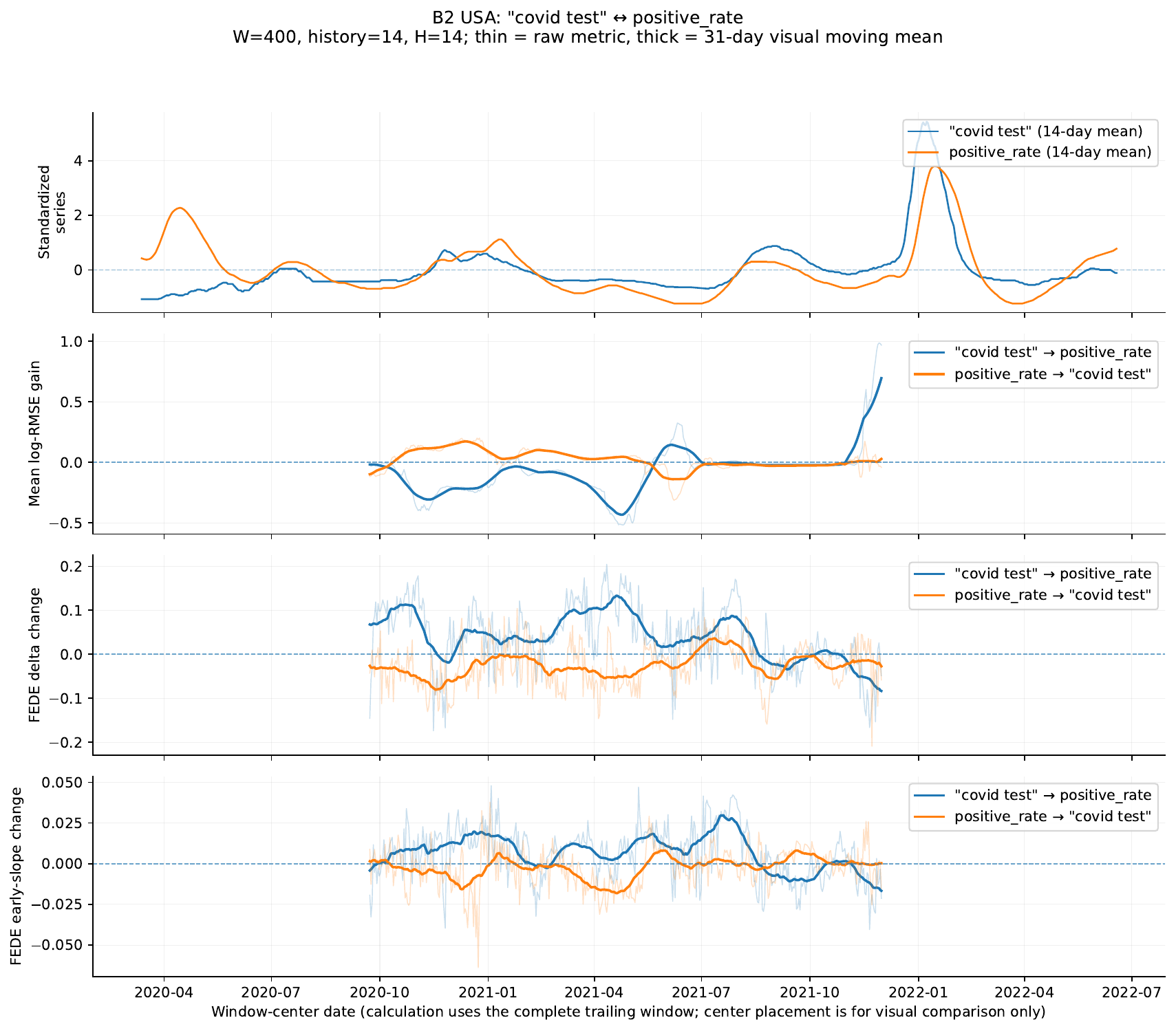}\\[-1mm]
\small (a) B2: ``covid test''$\leftrightarrow$positive rate
\end{minipage}\hfill
\begin{minipage}[t]{0.495\textwidth}\centering
\includegraphics[width=\linewidth]{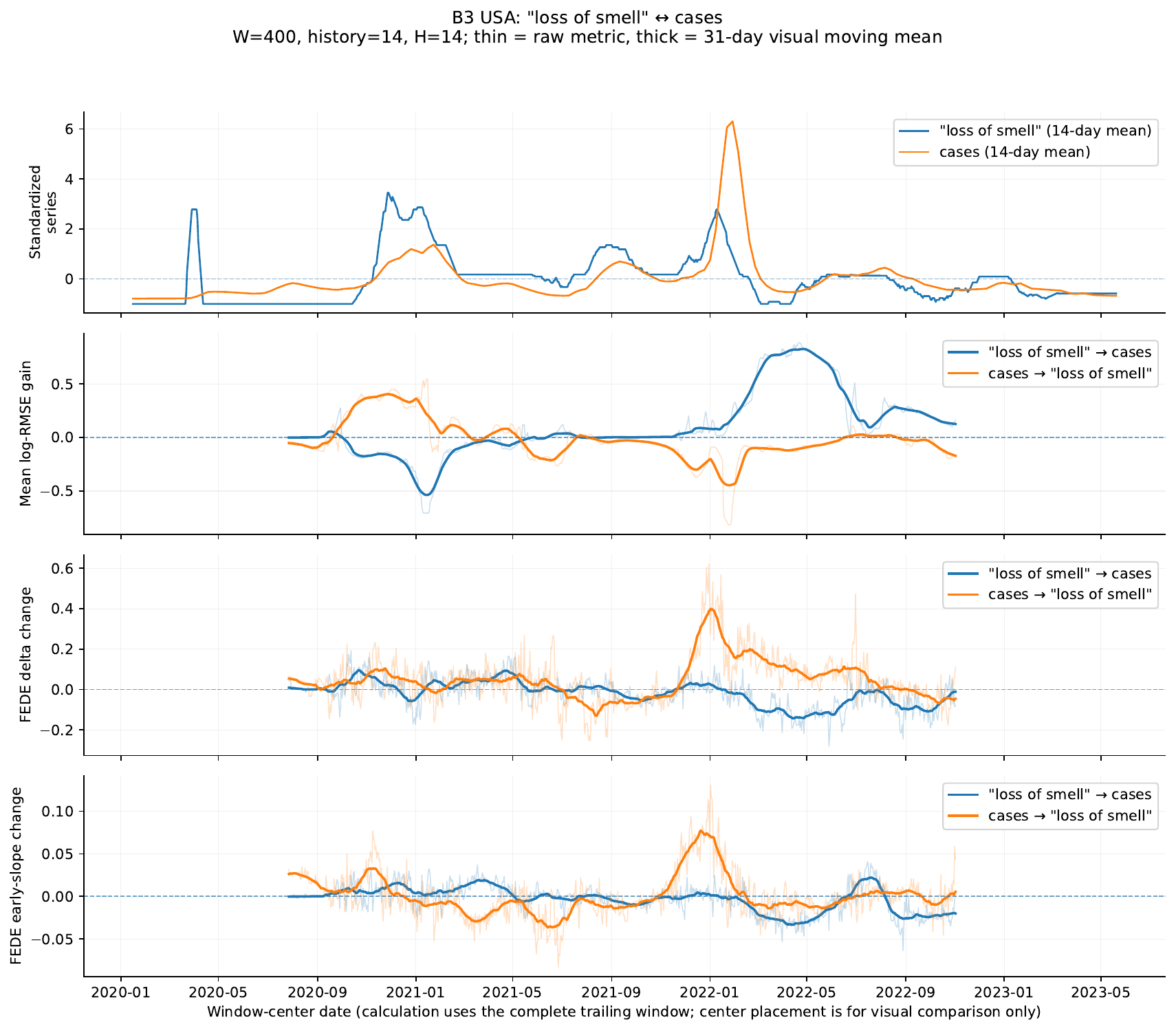}\\[-1mm]
\small (b) B3: ``loss of smell''$\leftrightarrow$cases
\end{minipage}
\noindent\small\textbf{Figure \thefigure.} Full rolling context plots for B2 and B3.\par\medskip
\normalsize

\par\medskip
\refstepcounter{figure}\label{fig:context_b4_b5}
\noindent\begin{minipage}[t]{0.495\textwidth}\centering
\includegraphics[width=\linewidth]{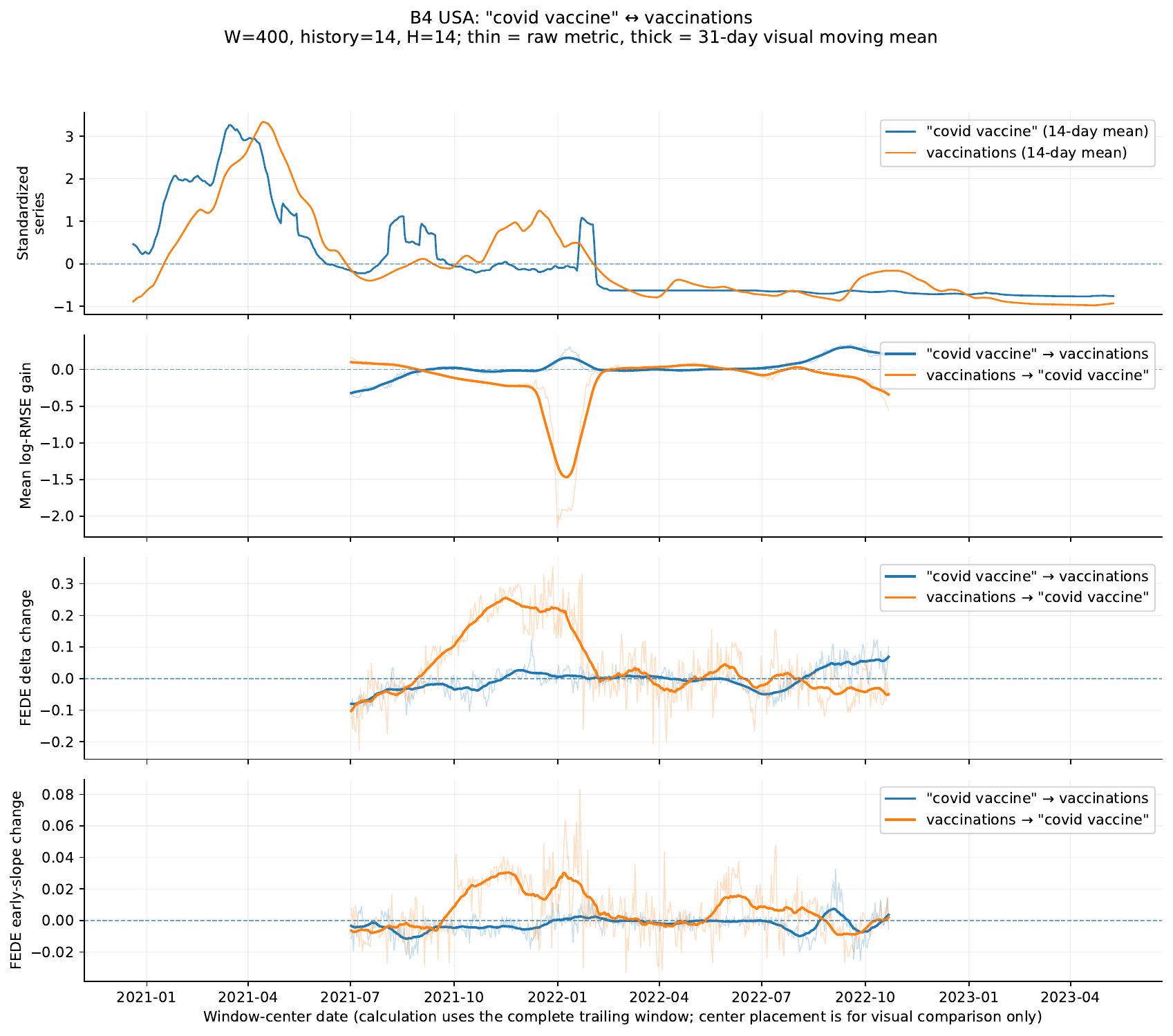}\\[-1mm]
\small (a) B4: ``covid vaccine''$\leftrightarrow$vaccinations
\end{minipage}\hfill
\begin{minipage}[t]{0.495\textwidth}\centering
\includegraphics[width=\linewidth]{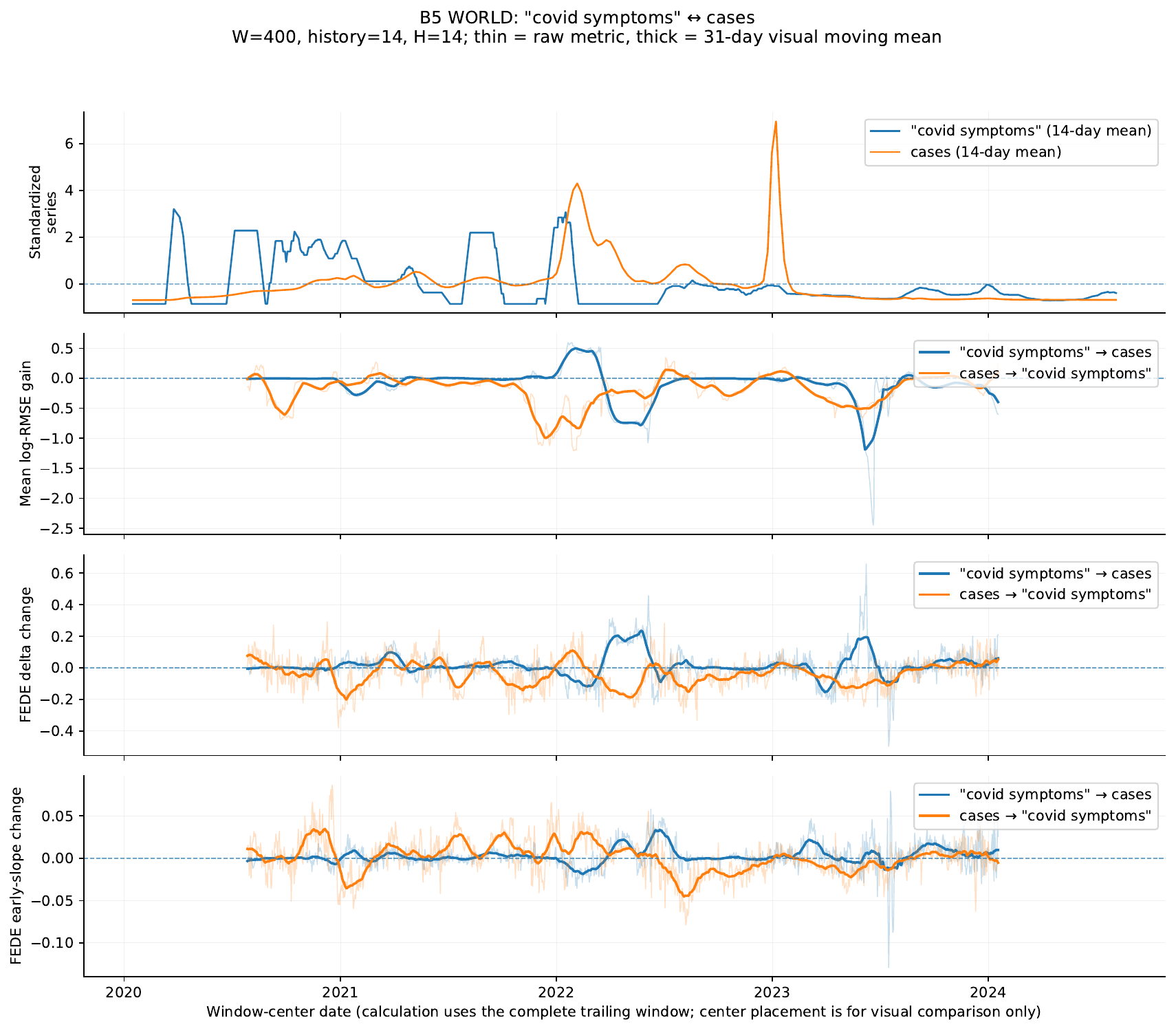}\\[-1mm]
\small (b) B5: world ``covid symptoms''$\leftrightarrow$cases
\end{minipage}
\noindent\small\textbf{Figure \thefigure.} Full rolling context plots for B4 and the world-level B5 pair.\par\medskip
\normalsize

\par\medskip
\refstepcounter{figure}\label{fig:context_b6_b7}
\noindent\begin{minipage}[t]{0.495\textwidth}\centering
\includegraphics[width=\linewidth]{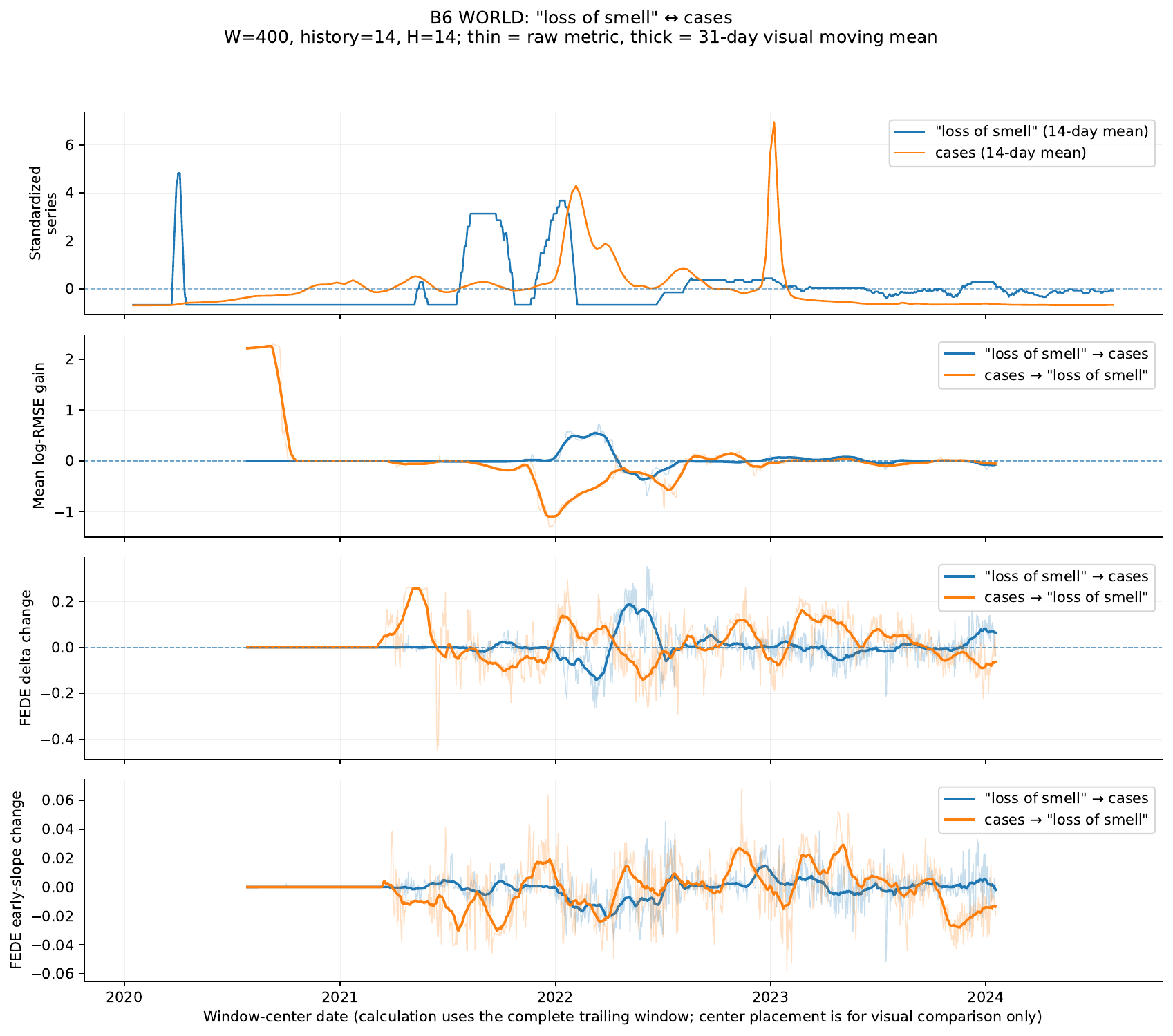}\\[-1mm]
\small (a) B6: world ``loss of smell''$\leftrightarrow$cases
\end{minipage}\hfill
\begin{minipage}[t]{0.495\textwidth}\centering
\includegraphics[width=\linewidth]{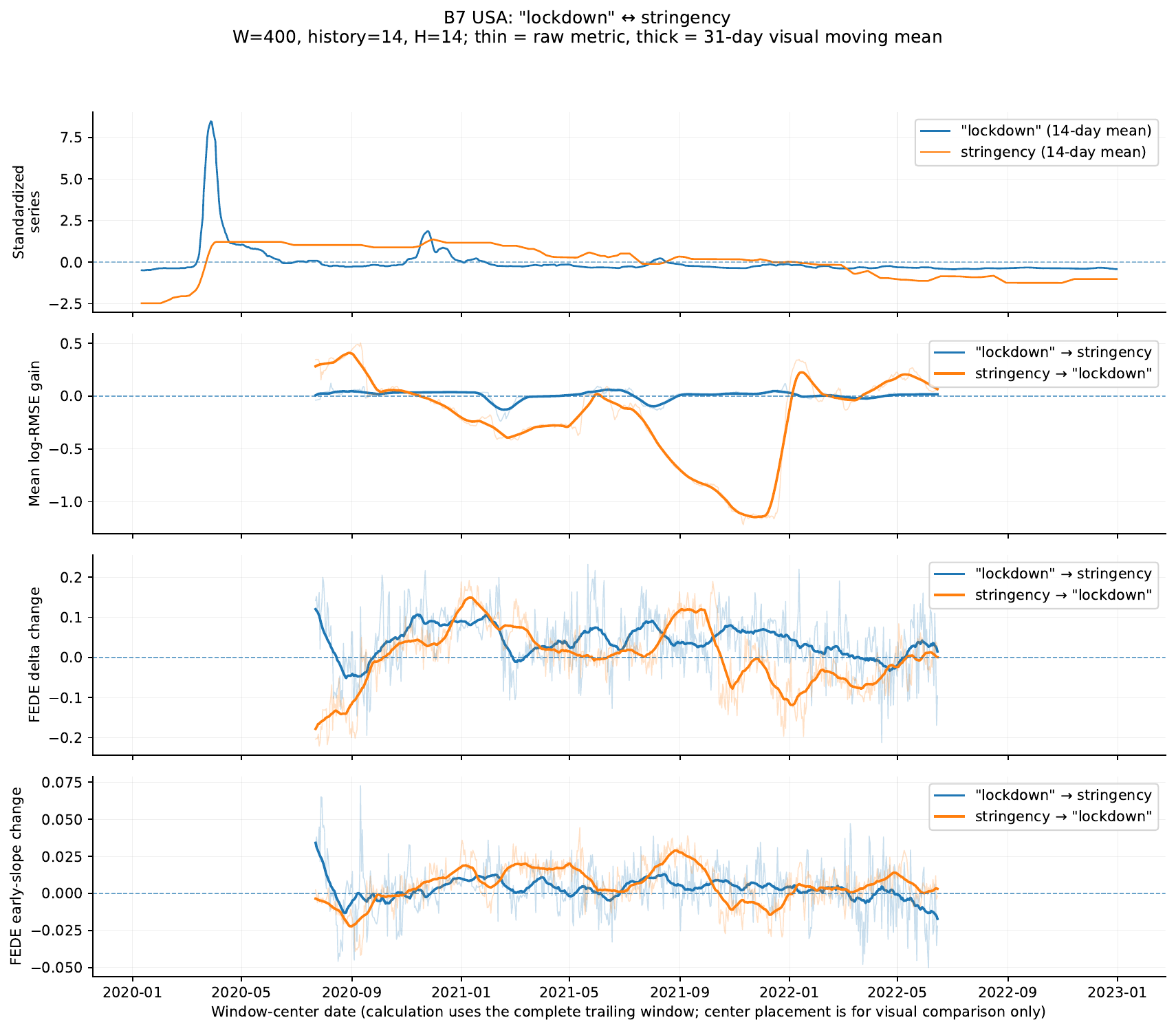}\\[-1mm]
\small (b) B7: ``lockdown''$\leftrightarrow$stringency
\end{minipage}
\noindent\small\textbf{Figure \thefigure.} Full rolling context plots for B6 and B7.\par\medskip
\normalsize

\par\medskip
\refstepcounter{figure}\label{fig:context_b8}
\begin{center}
\begin{minipage}[t]{0.66\textwidth}\centering
\includegraphics[width=\linewidth]{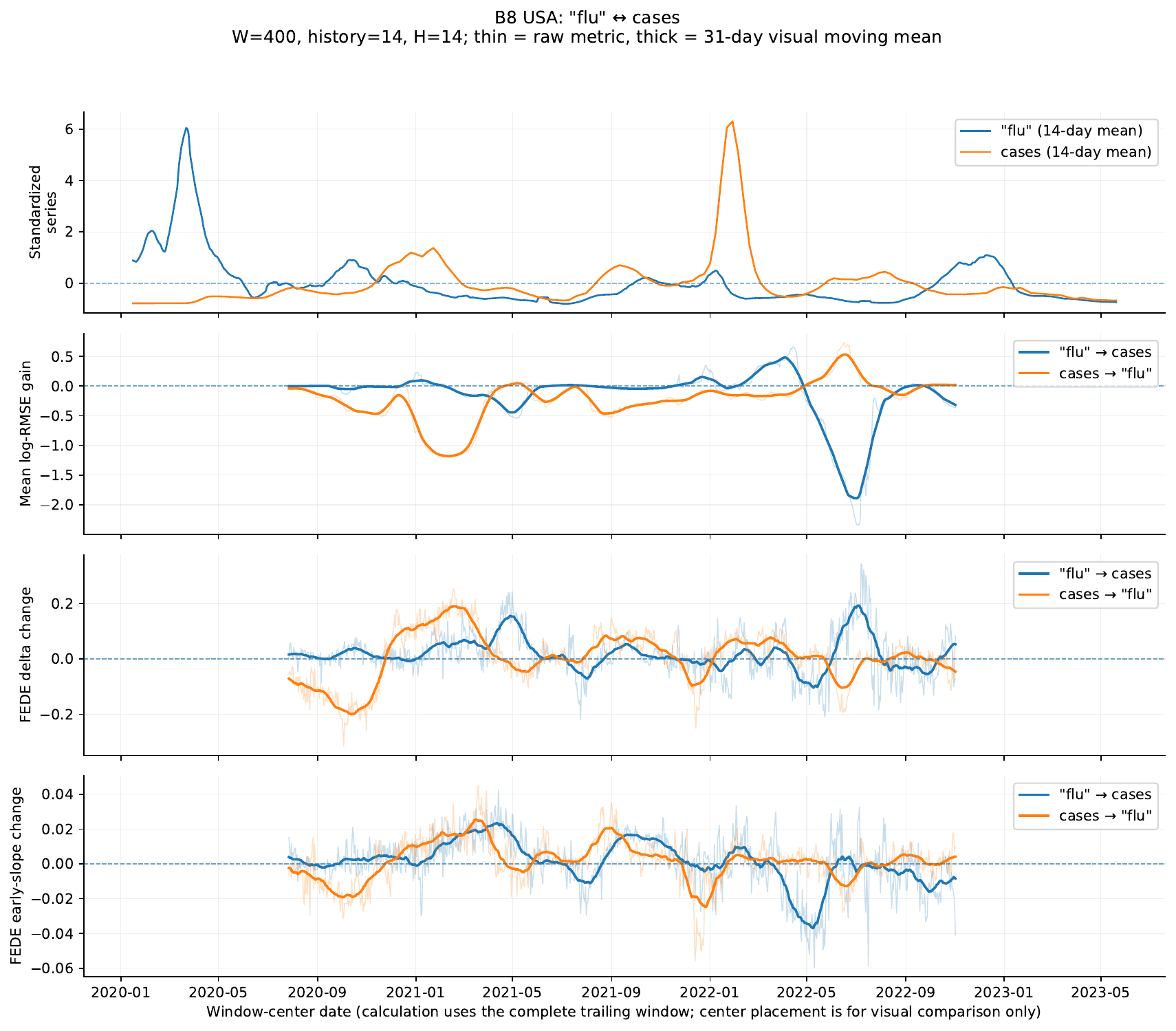}\\[-1mm]
\small B8: ``flu''$\leftrightarrow$cases
\end{minipage}
\end{center}
\noindent\small\textbf{Figure \thefigure.} Full rolling context plot for the seasonal/nonspecific stress pair B8.\par\medskip
\normalsize

The article-specific reproducibility archive is deposited on Zenodo \citep{VelichkoGoogleCOVID2026}, DOI \href{https://doi.org/10.5281/zenodo.23499353}{10.5281/zenodo.23499353}. It includes: (i) the Google Trends acquisition script and its acquisition settings; (ii) OWID download/aggregation scripts and the frozen compact source tables actually used in the analysis; (iii) a manifest with file hashes and source-date coverage; (iv) the pre-specified B1--B8 pair table; (v) the sequential $W=200,m=5$, $W=200,m=14$, and final $W=400,m=14,H=14$ configuration files; (vi) the postprocessing code used to obtain directional summaries, horizon profiles, and contextual plots; (vii) the directional-signature/DPRS calculator and its verified reference CSV; (viii) the conceptual equal-RMSE/different-entropy figure generator; and (ix) compact reference result tables sufficient to verify the article-scale calculations. The archive was independently rerun with the supplied 25-worker launcher, reproducing the final directional-signature table used in this manuscript.

\clearpage
\bibliographystyle{plainnat}
\bibliography{references}

\end{document}